\documentclass[journal]{IEEEtran}

\usepackage{times}
\usepackage{epsfig}
\usepackage{graphicx}
\usepackage{amsmath}
\usepackage{bbding}
\usepackage{pifont}
\usepackage{wasysym}
\usepackage{amssymb}
\usepackage{subfigure}
\usepackage{booktabs}
\usepackage{multirow}
\usepackage{algorithmicx,algorithm}
\usepackage{mathrsfs}
\usepackage{verbatim}
\usepackage{comment}
\usepackage{color}
\usepackage[noend]{algpseudocode}

\begin{document}
%
\title{Consistent Multiple Graph Embedding for Multi-View Clustering}

%
%

\author{Yiming~Wang,~Dongxia~Chang,~Zhiqiang~Fu
        and~Yao~Zhao,~\IEEEmembership{Senior~Member,~IEEE}
\thanks{Y. Wang, D. Chang, Z. Fu and Y. Zhao are with the Institute of Information
Science, Beijing Jiaotong University, Beijing 100044, China, and also
with Beijing Key Laboratory of Advanced Information Science and
Network Technology, Beijing 100044, China (e-mail: wangym@bjtu.edu.cn; dxchang@bjtu.edu.cn zhiqiangfu@bjtu.edu.cn;yzhao@bjtu.edu.cn).}}

%
%

\markboth{Journal of \LaTeX\ Class Files,~Vol.~14, No.~8, August~2015}%
{Shell \MakeLowercase{\textit{et al.}}: Bare Demo of IEEEtran.cls for IEEE Journals}
%



\maketitle

\begin{abstract}

Graph-based multi-view clustering aiming to obtain a partition of data across multiple views, has received considerable attention in recent years. Although great efforts have been made for graph-based multi-view clustering, it is still challenging to fuse characteristics from various views to learn a common representation for clustering. In this paper, we propose a novel Consistent Multiple Graph Embedding Clustering framework (CMGEC). Specifically, a multiple graph auto-encoder (M-GAE) is designed to flexibly encode the complementary information of multi-view data using a multi-graph attention fusion encoder. To guide the learned common representation maintaining the similarity of the neighboring characteristics in each view, a Multi-view Mutual Information Maximization module (MMIM) is introduced. Furthermore, a graph fusion network (GFN) is devised to explore the relationship among graphs from different views and provide a common consensus graph needed in M-GAE. By jointly training these models, the common representation can be obtained, which encodes more complementary information from multiple views and depicts data more comprehensively. Experiments on three types of multi-view datasets demonstrate CMGEC outperforms the state-of-the-art clustering methods.

\end{abstract}


%
\IEEEpeerreviewmaketitle

\section{Introduction}
%
%
%
%
\IEEEPARstart{W}{ith} the advance of information technology, multiple views of objects can be readily acquired in many domains. For instance, a piece of news can be reported by multiple news organizations, and an image can be described in different features: GIST, SIFT, and HOG, etc. Multi-view data can provide more comprehensive characteristics and helpful information than single-view \cite{pr/GaoMGW18,8896047}. With the advent of multi-view data, many multi-view clustering methods \cite{TAI/SURVEY} have emerged and are widely applied in medicine \cite{isci/ChaoSLWLLB19}, and computer vision \cite{eccv/RiemenschneiderBWG14}, etc. For example, Chao {\em et al.} \cite{isci/ChaoSLWLLB19} propose a multi-view co-clustering algorithm and apply the algorithm to an opioid dependence treatment study. However, there are still some challenges to multi-view clustering. Large differences between data from different views may produce view disagreement, which can distort a similarity matrix used to depict samples within the same class \cite{tnn/YinGXG19,tip/WangLWZZH15}. Additionally, the dimension difference of different features can lead to difficulties in feature fusion \cite{tbd/Zheng15}.

Various multi-view clustering methods have been proposed to solve the above problems, including graph-based methods\cite{ijcai/NieLL17, kbs/KangSHCPZX20}, subspace-based methods \cite{pami/ZhangFHCXTX20}, kernel-based methods \cite{ijcai/WangLZTLHXY19, pami/LiuZLWTYSWG19}, etc. The graph-based multi-view clustering methods \cite{tkde/ZhanNCNZY19,9154578} seek to find a fusion graph across all views and use graph-cut algorithms or other technologies to produce the clustering assignments. This kind of method can solve the problem of dimensional differences of different views. However, graph-based multi-view clustering methods are generally shallow models with limited capacity to reveal the relations in complex multi-view data. Moreover, these models can hardly combine graph structural information with data intrinsic characteristics, which are equally important for clustering tasks. 

Graph Convolutional Networks (GCN) \cite{iclr/KipfW17,esws/SchlichtkrullKB18} recently emerged can encode both the graph structure and node characteristics for latent node representation. The GCN follows a message-passing manner that aggregates a node feature information from its topological neighbors in each convolutional layer. It is consistent with the clustering task of aggregating similar samples into clusters. Thus, many GCN-based clustering methods \cite{ijcai/WangPHLJZ19,cvpr/Wang0LW19} have been proposed. For instance, to jointly integrate the information from node content and consensus graph, \cite{ijcai/TaoLLW019} employs graph representation learning techniques to ensemble clustering. To achieve mutual benefit for both learned embedding and graph clustering, Wang {\em et al.} \cite{ijcai/WangPHLJZ19} propose a goal-directed graph attentional autoencoder for attributed graph clustering. These GCN-based methods greatly improve the performance of graph-based clustering. However, all these methods can handle single view data, and there are few GCN-based multi-view clustering algorithms. To model multi-view graph information, Fan {\em et al.} \cite{www/FanWSLLW20} employ one informative graph view to reconstruct multiple graph views to capture the shared representation of multiple graphs. Unlike the general multi-view clustering method applied to multi-view data and graphs, it is employed for single-view data with multi-view graphs. The above methods apply GCN to exploit both graph structure and node content to learn a latent representation. However, the graph in most GCN-based methods is fixed, making the clustering performance heavily dependent on the predefined graph. And a noisy graph with unreliable connections can result in ineffective convolution with wrong neighbors on the graph \cite{nips/YunJKKK19}, which may worsen graph clustering performance. 

In order to solve the above challenges in multi-view clustering, we propose a Consistent Multiple Graph Embedding Clustering framework (CMGEC), which is mainly composed of Multiple Graph Auto-Encoder (M-GAE), Multi-view Mutual Information Maximization module (MMIM) and Graph Fusion Network (GFN). Our major contributions can be summarized as follows:
\begin{itemize}
    \setlength{\itemsep}{0pt}
    \setlength{\parsep}{0pt}
    \item To capture the complementary information and internal relations of each view well, we propose a multi-graph attention fusion encoder to adaptively learn a common representation from multiple views. 
    \item To maintain consistency within views, multi-view mutual information maximization is devised to make similar instances still similar to each other in the common space.
    \item To explore the relationships among different view graphs, a graph fusion network is devised to fuse graphs from multiple views to get a consensus graph needed in the multiple graph auto-encoder. And to improve the separability of the consensus graph, the rank constraint on its Laplacian matrix is utilized to train the GFN.
    \item We have conducted experiments on three types of multi-view data, and experiments show that our CMGEC outperforms state-of-the-art clustering methods.
\end{itemize}

\begin{figure*}[!t]
	\centering
	\includegraphics[width=14cm]{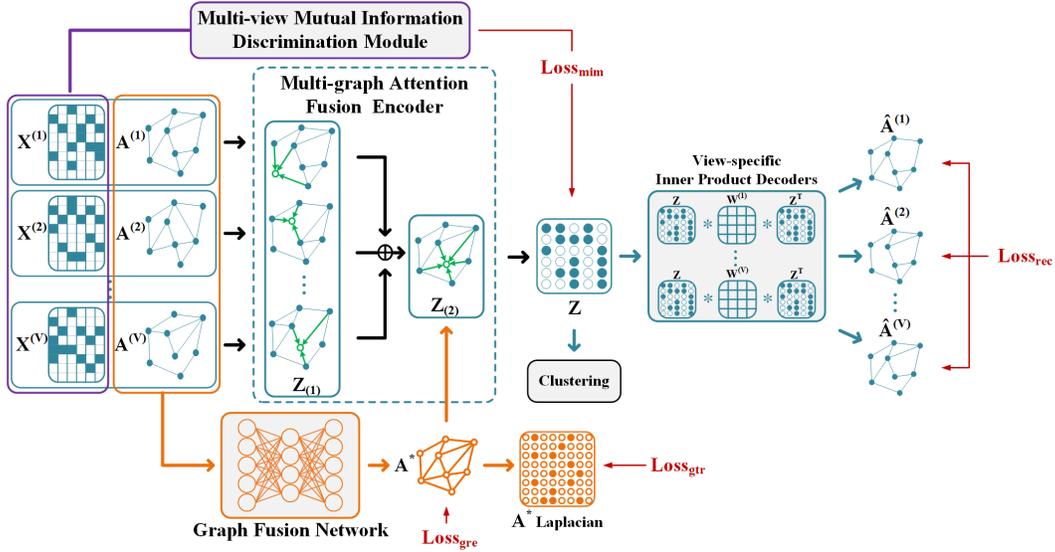}
	\caption{The framework of the proposed CMGEC. It consists of three main components: Multiple Graph Auto-Encoder(M-GAE), Multi-view Mutual Information Maximization module (MMIM), and Graph Fusion Network (GFN).}
\end{figure*}

\section{Related Works}
Before introducing the proposed CMGEC, multi-view clustering and mutual information maximization are briefly introduced in this section. 

\paragraph{Multi-view clustering}
Recently, multi-view learning has attracted lots of attention, and numerous multi-view clustering methods have been proposed. We roughly divide them into three categories: graph-based methods \cite{tkde/WangYL20}, kernel-based methods\cite{pami/LiuZLWTYSWG19,9447203}, and subspace-based methods \cite{pami/ZhangFHCXTX20}. Subspace-based multi-view clustering uses subspace learning to obtain a latent subspace shared by different views, which solves the difficulty of handling high-dimensional data to some extent. For example, to alleviate the problem that most subspace-based methods are heavily influenced by the original features, \cite{pami/ZhangFHCXTX20} explores underlying complementary information from multiple views and simultaneously seeks the underlying latent representation. Introducing neural networks, Li {\em et al.} \cite{iccv/LiZF0ZH19} construct subspace representations linked with a latent representation to identify the underlying cluster structure of high-dimensional data. Kernel-based methods typically combine a set of base kernels constructed in different views to obtain the clustering results. To alleviate the problem of the high computational complexity of kernel-based clustering, Wang {\em et al.} \cite{ijcai/WangLZTLHXY19} first maximize alignment between consensus clustering matrix and weighted base partitions. To further reduce storage and computational complexity, \cite{pami/LiuZLWTYSWG19} jointly learns a consensus clustering matrix, imputes each incomplete base matrix, and optimizes the corresponding permutation matrices. These multi-view methods show satisfactory performance but suffer from two main disadvantages: (a) They can typically employ one of graph structure or data characteristics; (b) Subspace-based multi-view methods are typically sensitive to initialization; (c) Kernel-based multi-view methods suffer from intensive computational complexities. 

For the graph-based multi-view clustering methods, the view-specific graphs are constructed based on the $k$-NN graph and used to find a fusion graph across all views. Most graph-based multi-view clustering methods are based on spectral clustering, a classic data clustering algorithm aiming to build a normalized affinity matrix and compute the eigenvectors of this normalized affinity matrix. Combined with graph fusion, it can be extended to multi-view clustering. Based on spectral partitioning and local refinement, Chikhi \cite{ipm/Chikhi16} presents a parameter-free multi-view spectral clustering algorithm. To address the issue that dependencies among views often delude correct predictions, Son et al. \cite{aaai/SonJLK17} propose a spectral clustering method to deal with multi-view data and dependencies among views based on the brainstorming process. Nie {\em et al.} \cite{ijcai/NieLL17} propose a Laplacian rank constrained graph, which can be approximate as the centroid of the built graph for each view with different confidences. In order to solve the problem that graph-based clustering highly depends on the quality of a predefined graph, \cite{tcyb/ZhanZGW18} learns a global graph, which has an exact number of the connected components that reflects cluster indicators. To sufficiently consider weights of different views, Wang {\em et al.} \cite{tkde/WangYL20} propose a graph-based multi-view clustering (GMC) method coupling the learning of the similarity-induced graphs, the unified graph, and the clustering task into a joint clustering framework. However, one major drawback of these shallow models is that they have limited capacity to reveal the deep relations in complex graph data.

\paragraph{Mutual information maximization} 
To maintain the consistency of similar samples in each view, we employ mutual information maximization in our model. Mutual information is a Shannon entropy-based fundamental quantity for measuring the relationship between random variables \cite{icml/BelghaziBROBHC18}. Following regularized information maximization (RIM) \cite{RIM}, maximizing the mutual information between input samples and latent cluster assignments can be used in discriminative clustering. Some deep clustering methods further study this concept \cite{icml/BelghaziBROBHC18, iclr/HjelmFLGBTB19}, which learn discriminative neural network classifiers that maximize the mutual information. Furthermore, mutual information is also widely used in multi-view learning. In particular, \cite{nips/BachmanHB19} proposes a self-supervised representation learning based on maximizing mutual information between features extracted from multiple views of a shared context. Mao {\em et al.} \cite{aaai/MaoYGY21} propose deep mutual information maximum (DMIM) for cross-modal clustering, which preserves the shared multi-view information while eliminating the superfluous information of individual modalities in an end-to-end manner. Generally, mutual information maximization corresponds to maximizing the following objective:
\begin{equation}
I(X,K) = H(K)-H(K|X)
\end{equation}
where $H(\cdot)$ and $H(\cdot|\cdot)$ are the entropy and conditional entropy, respectively. $K\in \{1,...,\mathbb{K}\}$ and $X\in \mathbb{X}$ denote random variables for cluster assignments and data samples, respectively. 
And the mutual information between sample $X$ and latent representation $Z$ can be understood as:
\begin{equation}
\begin{aligned}
I(X,Z)&=  \int\!\!\!\int p(z|x)p(x)\log\frac{p(z|x)}{p(z)}\text{d}x\text{d}z\\
      &=  KL(p(z|x)p(x)||p(z)p(x))\\
\end{aligned}
\end{equation}
where $p(x)$ is the distribution of the input samples and $p(z|x)$ is the distribution of the latent representations. The distribution of the latent space $p(z)$ can be calculated by $p(z)=\int\! p(z|x)p(x)\text{d}x$. And adversarial learning can be used to constrain the latent representations to have desired statistical characteristics specific to the input samples.

\section{The Proposed Model}

As aforementioned, current graph-based multi-view clustering methods have the following shortcomings: (a) The shallow model can hardly combine the graph structural information with the node intrinsic characteristics; (b) The GCN-based methods generally use a fixed graph structure, and its performance depends heavily on the predefined graph. To handle these two challenges, CMGEC is proposed, and the overall framework is shown in Fig.~1. Our CMGEC mainly contains three parts: M-GAE, MMIM, and GFN. Firstly, the predefined graph of each view is input into the GFN to obtain the consensus graph. In order to make the consensus graph more sparse and cluster-friendly, the rank constraint on its Laplacian matrix is used to train the GFN. Then the raw features, the graph of each view, and the consensus graph are fed into the M-GAE to learn a common latent representation. To flexibly incorporate information from all views, a multi-graph attention fusion encoder is introduced into the M-GAE. Moreover, MMIM is devised to make the learned common representation maintain the similarity of the neighboring characteristics. In the following, we will describe our proposed model in detail.

Formally, given a multi-view dataset $\mathcal{X} = \{X^{(v)}\}_{v=1}^V$, consisting of $N$ samples from $V$ views, $ X^{(v)}\in {R}^{ N \times d_v}$ denotes the feature matrix of the $v$-th view. $d_v$ is the dimension of the feature of the $v$-th view. $\mathcal{A} = \{A^{(v)}\}_{v=1}^V \in \mathbb{R}^{N \times N}$ represents the graph of each view. $A^*$ denotes the common graph, where $A^* = f(\mathcal{A};\theta_g) \in \mathbb{R}^{N \times N} $ is learned by the GFN. The parameters of the GFN are defined by $\theta_g$. $Z$ denotes the common latent representation, where $Z = f(\mathcal{X},\mathcal{A},A^*;\theta_e) \in \mathbb{R}^{N \times m} $  is learned by the multiple graph fusion encoder $E$. The parameters of the encoder are defined by $\theta_e$, and $m$ is the dimension of the learned common representation. $\{\hat{A}^{(v)}\}_{v=1}^V = f(Z;\theta_d)\in \mathbb{R}^{m \times m}$ represents the reconstructed graph relation, which is the output of the view-specific decoders $D$, and the parameters of the decoders are denoted by $\theta_d$.

\subsection{Multiple Graph Auto-Encoder}

In order to represent both multi-view graph structure and node feature comprehensively in a unified framework, we develop an M-GAE in which a multi-graph attention fusion encoder learns common latent representation. Moreover, the view-specific decoders are designed to reconstruct multi-view graph data from the learned representation.

\paragraph{Multi-Graph Attention Fusion Encoder} To learn a common representation that can fully integrate information from multiple views, a multi-graph attention fusion layer, which can fuse multi-view data and graphs adaptively, is devised based on the GCN \cite{iclr/KipfW17}. After getting the common representation, the final representation $Z$ can be obtained by the GCN using the common representation and graph. Here, the common graph is obtained by GFN, and we will introduce it in detail in the next section.

The GCN extends the operation of convolution to graph data in the spectral domain. Here, $Z_{(l)}$ is the representation learned by the $l$-th layer of GCN, and it can be obtained by the following graph convolutional operation:
\begin{equation}
Z_{(l)} = \phi\left(\tilde{D}^{-\frac{1}{2}}\tilde{A}\tilde{D}^{-\frac{1}{2}}Z_{(l-1)}W_{(l)}\right)
\end{equation}
where $\tilde{A} = A+I$ and $\tilde{D}_{ii} = \sum_j\tilde{A}_{ij}$. $I$ is the identity diagonal matrix, $W_{(l)}$ denotes the learned parameter matrix, and $\phi(\cdot)$ is an activation function. 

In the multi-graph attention fusion encoder, the first layer is composed of $v$ view-specific GCN layers, and the input of the first layer are multi-view data $\mathcal{X} = \{X^{(v)}\}_{v=1}^V$ and the corresponding graphs $\mathcal{A} = \{A^{(v)}\}_{v=1}^V$. Then, the $v$-th view-specific representations $Z^{(v)}_{(1)}$ learned by the first layer can be obtained by:
\begin{equation}
Z^{(v)}_{(1)} = \phi\left((\tilde{D}^{(v)})^{-\frac{1}{2}}\tilde{A}^{(v)}(\tilde{D}^{(v)})^{-\frac{1}{2}}X^{(v)}W^{(v)}_{(1)}\right)
\end{equation}
In order to flexibly integrate the view-specific representation, a multi-graph attention fusion layer is devised using the attention strategy. Unlike graph attention networks that learn hidden representations of each node by weighting the representations for each nearest neighbor in the same view, our approach focuses on the weighting of different views. To adaptively fuse the representation of a sample in different views, attention coefficient matrix $W_a$ is introduced to learn the importance of different views. Hence, the common representation learned by the multi-graph attention fusion layer can be obtained by the following operation:
\begin{equation}
Z_{(2)}= \phi\left(\sum_{v=1}^VW_{a}\left((\tilde{D}^{(v)})^{-\frac{1}{2}}\tilde{A}^{(v)}(\tilde{D}^{(v)})^{-\frac{1}{2}}Z^{(v)}_{(1)}W^{(v)}_{(2)}\right)\right)
\end{equation}
Then a GCN layer is used to provide the final common representation $Z$:
\begin{equation}
Z = \phi\left((\tilde{D}^*)^{-\frac{1}{2}}\tilde{A}^*(\tilde{D}^*)^{-\frac{1}{2}}Z_{(2)}W_{(3)}\right)
\end{equation}
where $A^*$ is the consensus graph learned by GFN, and $\tilde{D}^*_{ii} = \sum_j\tilde{A}^*_{ij}$.

\paragraph{View-specific Graph Decoders} In order to guide the multi-graph fusion encoder to learn a comprehensive common representation, view-specific graph decoders are applied to reconstruct the multi-view graph data $\hat{A}^{(1)},...,\hat{A}^{(v)}$ from the learned representation $Z$. As the learned representation already contains both contents and structure information, inner product decoders are adopted to predict the links between nodes, which can be written as:
\begin{equation}
\hat{A}^{(v)} = {\rm sigmoid} (Z\cdot W^{(v)}\cdot Z^T) 
\end{equation}
where $W^{(v)}$ is the learned parameter matrix in the $v$-th view-specific decoder.

\paragraph{Reconstruction loss} To train the M-GAE, we minimize the sum of reconstruction error of each view by measuring the difference between $A^{(v)}$ and $\hat{A}^{(v)}$:
\begin{equation}
L_{rec} = \sum_{v=1}^V L_{rec}^{(v)} = \sum_{v=1}^V loss(A^{(v)},\hat{A}^{(v)})
\end{equation}
where $L_{rec}^{(v)}$ is the reconstruction loss for the $v$-th view and $L_{rec}$ is the reconstruction loss for all views. 

\subsection{Graph Fusion Network}
Multi-view data and graphs provide multiple independent and complementary information from multiple feature spaces, and their analysis can often result in more integrated and accurate results than single view \cite{tcss/LiL18}. However, each graph contains different adjacency relations in different views and cannot be used directly in the common space. Therefore, to explore relationships among different views and provide global node relationships, a graph fusion network is devised to produce the consensus graph $A^*$.

In our model, a fully connected network is employed to learn the consensus graph. Specifically, the consensus graph learned by the $l$-th layer in the graph fusion network can be described as:
\begin{equation}
G_{(l)} = \phi(W_{g(l)}G_{(l-1)}+b_{g(l)})
\end{equation}
where $\phi$ is the activation function of the fully connected layers, $W_{g(l)}$ and $b_{g(l)}$ are the weight matrix and bias of the $l$-th layer in the graph fusion network, respectively. To adaptively fuse each graph, a multi-graph fusion layer is placed on the first layer of GFN, and it is defined as:

\begin{equation}
G_{(1)}= \phi(\sum_{v=1}^VW_{f}(W^{(v)}_{g(1)}A^{(v)}+b^{(v)}_{g(1)}))
\end{equation}
where $W_f$ is the attention coefficient matrix that indicates the importance of the edge in different views.

In order to combine the features of each graph and make the consensus graph $A^*$ more suitable for clustering, the loss of the graph fusion network is defined as
\begin{equation}
\begin{aligned}
 L_G = & L_{gre} +\lambda_1 L_{gtr} \\
 = &\sum_{v=1}^V loss(A^{(v)},A^*) + \lambda_1 tr(Q^TL_{A^*}Q) \\  
 &s. t. \  Q^TQ = I 
\end{aligned}
\end{equation}
where $L_{gre}$ and $L_{gtr}$ are the graph reconstruction loss for all views and the Ratio Cut used spectral clustering, respectively. $tr(\cdot)$ is the trace operator. $L_{A^*}$ is the Laplacian matrix of $A^*$ and $Q$ is the relaxed indicator which can be computed by the eigenvalue decomposition of $L_{A^*}$. $\lambda_1$ is a hyperparameter that balances these two losses.

Obviously, the consensus graph $A^*$ can be segmented directly to obtain the clustering results. Since the GFN uses only the graph structure and ignores the intrinsic characteristics of nodes, the clustering result obtained using $A^*$ is worse than that of the learned common representation.

\subsection{Multi-view Mutual Information Maximization}

\begin{figure}[t]
	\centering
	\includegraphics[width=8.8cm]{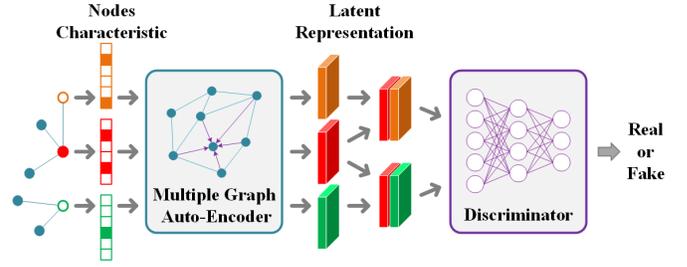}
	\caption{The architecture of MMIM. In the training process, we select $k_M$ nearest neighbors for each nodes in each view. And the positive pairs are composed of the common representation of nodes and their nearest neighbors. Meanwhile, the same number of nodes besides the nearest neighbors are randomly selected. And the negative pairs consist of the common representation of nodes and these random nodes. Finally, a discriminator is used to distinguish these pairs.}
\end{figure} 

In the fields of subspace learning, it has been well recognized that if two samples are close to each other, their corresponding low dimensional representations should also be close in the latent space \cite{tcyb/WenHFFYZ19}. Therefore, the inter-sample information can be used to guide the autoencoder to learn more cluster-friendly representation.

In our model, we assume that if two samples $x$ and $x'$ are close in any view, their corresponding representations $z$ and $z'$ should also be close in the common latent view. Based on the assumption, a multi-view mutual information maximization module(MMIM) is devised to boost the similarity of inter-neighbour representations. The mutual information maximization module is shown in Fig.~2.

Since larger mutual information denotes the representations are more similar, the mutual information is expected to be as large as possible, and the objective of MMIM can be described as
\begin{equation}
\max\{I(X,Z')\}
\end{equation}
According to Eq.(2) and Eq.(12), the loss function of MMIM $L_{mim}$ can be written as 
\begin{equation}
L_{mim} = -KL(p(z'|x)p(x)||p(z')p(x))
\end{equation}
However, KL divergence is unbounded. Therefore, we use JS divergence instead of KL divergence in mutual information and Eq.(13) can be converted to
\begin{equation}
L_{mim} = -JS(p(z'|x)p(x)||p(z')p(x))
\end{equation}
According to \cite{kt}, the variational estimation of JS divergence between two distributions $p(x)$ and $q(x)$ is defined as
\begin{equation}
\begin{aligned}
JS(p(x)||q(x))= & \mathbb{E}_{x\sim p(x)}[\log \rho(T(x))]\\
              + & \mathbb{E}_{x\sim q(x)}[\log (1-\rho(T(x)))]\\
\end{aligned}
\end{equation}
where $T(x)=\log \frac{2p(x)}{p(x)+q(x)}$ \cite{kt}. In our loss function,  $p(z'|x)p(x)$ and $p(z')p(x)$  are used to replace $p(x)$ and $q(x)$. Hence, substituting Eq.(15) into Eq.(14) yields 
\begin{equation}
\begin{aligned}
&&L_{mim} = &- \mathbb{E}_{(x,z')\sim p(z'|x)p(x)}[\log \rho(T(x,z'))]&\\
&&\         &- \mathbb{E}_{(x,z')\sim p(z')p(x)}[\log(1-\rho(T(x,z')))]&
\end{aligned}
\end{equation}

Here, negative sample estimation \cite{iclr/HjelmFLGBTB19} is used to solve the problem in Eq.(16). Positive and negative sample pairs are generated by the latent representations. Then a discriminator is used to distinguish the negative sample pairs and positive sample pairs to estimate the distribution of positive samples. In Eq.(16), $\rho(T(x,z'))$ is a discriminator, and the representation of sample $x$ and its nearest neighbors $x'$ compose positive pairs. The negative pairs are composed of the representation of $x$ and random representations outside the nearest neighbors. For each sample, $k_M$ nearest neighbors are selected to compose positive pairs in all views. For data without a graph, we use $k$-NN algorithm to find the nearest neighbors. And for data with attributed graph, we use a modified Shared Nearest Neighbor(SNN) \cite{jarvis1973clustering} similarity to find the nearest neighbors on the attributed graph. The modified SNN similarity can be written as
\begin{equation}
sim(i,j)=\left\{\begin{matrix}
0 ,& No\ edge\ between\ v_i\ and\ v_j,\\ 
|\mathcal{N}(i)\cap \mathcal{N}(j)|, & v_i\ and\ v_j\ are\ adjacent.
\end{matrix}\right.
\end{equation}
where $\mathcal{N}(*)$ denotes the neighboring nodes of $v_*$. After getting the similarity of all other points to $v_*$, the $k_M$ most similar nodes are selected as the nearest neighbors of $v_*$.

Thus the total objective function of M-GAE module is defined as:
\begin{equation}
L_{M} = L_{rec}+ \lambda_2 L_{mim}
\end{equation}
where $\lambda_2$ is a hyperparameter that balances these two loss functions.
\section{Experiments}


\begin{table}[!t]
\centering
\caption{Datasets statistics of multi-view data without predefined attribute graph.}
\resizebox{8.5cm}{!}{
\begin{tabular}{lcccccc}
\toprule
Datasets   & Classes & Nodes  & Feature1  & Feature2  & Feature3  & Feature4 \\
\midrule
3source    & 6       & 169        &3560 &3631 &3068 &-        \\
BBC        & 5       & 685        &4659 &4633 &4665 &4684    \\
100LEAVES  & 100     & 1600       &64   &64   &64   &-       \\
Cub        & 10      & 600        &1024 &300  &-    &-       \\
\bottomrule
\end{tabular}}\label{tab:data1}
\end{table}

\begin{table}[!t]
\centering
\caption{Datasets statistics of multi-view data with common attribute graph.}
\resizebox{8.5cm}{!}{
\begin{tabular}{lccccc}
\toprule
Datasets   & Classes & Nodes &Edges   & Attribute1  & Attribute2   \\
\midrule
Cora       & 7       & 2708  & 5429   &1433 &2708   \\
Citeseer   & 6       & 3327  & 4732   &3703 &3327   \\
Pubmed     & 3       & 19717 & 44438  &19717&500      \\
\bottomrule
\end{tabular}}\label{tab:data2}
\end{table}

\begin{table}[!t]
\centering
\caption{Datasets statistics of single-view data with multiply attribute graphs.}
\resizebox{8.5cm}{!}{
\begin{tabular}{lcccccc}
\toprule
Datasets   & Classes & Nodes  &Dimension  & Edge1  & Edge2 & Edge3   \\
\midrule 
DBLP       & 4       & 4057   & 334       &11113 &5000495&6776335   \\
IMDB       & 3       & 4780   & 1232      &98010 &21018&-   \\
ACM        & 3       & 3025   & 1830      &29281&2210761 &-     \\
\bottomrule
\end{tabular}}\label{tab:data3}
\end{table}

\begin{table*}[!ht]
    \small
	\centering
	\caption{Performance comparisons between single view methods and CMGEC on multi-view data without predefined attribute graph.}
	\resizebox{12.5cm}{!}{
	\begin{tabular}{llccccc}
		\toprule
		Datasets & Methods & ACC & NMI  & ARI & AMI & F1  \\ \midrule
		\multirow{5}{*}{3Sources}     
		& KM++ \cite{www/Sculley10}  & 0.5390$\pm$0.0647 & 0.4320$\pm$0.1035 & 0.2880$\pm$0.1300 & 0.4000$\pm$0.1094 & 0.3260$\pm$0.0815 \\
		& GAE \cite{corr/KipfW16a}        & 0.6765$\pm$0.0155 & 0.5756$\pm$0.0476 & 0.4553$\pm$0.0446 & 0.5535$\pm$0.0499 & 0.6317$\pm$0.0136 \\
		& DAEGC \cite{ijcai/WangPHLJZ19} & \textcolor{blue}{0.7160$\pm$0.0000} & \textcolor{blue}{0.6066$\pm$0.0000} & \textcolor{red}{0.6208$\pm$0.0000} & \textcolor{blue}{0.6135$\pm$0.0000} & \textcolor{blue}{0.6470$\pm$0.0000} \\
		& SDCN \cite{www/Bo0SZL020}       & 0.6252$\pm$0.0028 & 0.4230$\pm$0.0088 & 0.4225$\pm$0.0143 & 0.3921$\pm$0.0096 & 0.3487$\pm$0.0032 \\
		& CMGEC     & \textcolor{red}{0.7653$\pm$0.0307} & \textcolor{red}{0.6694$\pm$0.0143} & \textcolor{blue}{0.6049$\pm$0.0446} & \textcolor{red}{0.6515$\pm$0.0147} & \textcolor{red}{0.6634$\pm$0.0355} \\ 
		\midrule
		\multirow{5}{*}{BBC}    
		& KM++ \cite{www/Sculley10}  & 0.5201$\pm$0.0673 & 0.3516$\pm$0.0732 & 0.2180$\pm$0.1006 & 0.3458$\pm$0.0745 & 0.4147$\pm$0.0764 \\
		& GAE \cite{corr/KipfW16a}        & 0.6397$\pm$0.0066 & 0.5265$\pm$0.0291 & 0.4720$\pm$0.0437 & 0.5228$\pm$0.0292 & \textcolor{blue}{0.6260$\pm$0.0142} \\
		& DAEGC \cite{ijcai/WangPHLJZ19} & 0.6746$\pm$0.0000 & 0.5278$\pm$0.0000 & 0.4661$\pm$0.0000 & 0.5121$\pm$0.0000 & 0.6606$\pm$0.0001 \\
		& SDCN \cite{www/Bo0SZL020}       & \textcolor{blue}{0.7156$\pm$0.0044} & \textcolor{blue}{0.5713$\pm$0.0020} & \textcolor{blue}{0.5082$\pm$0.0067} & \textcolor{blue}{0.5664$\pm$0.0020} & 0.4775$\pm$0.0020 \\
		& CMGEC    & \textcolor{red}{0.8737$\pm$0.0061} & \textcolor{red}{0.7144$\pm$0.0119} & \textcolor{red}{0.7392$\pm$0.0115} & \textcolor{red}{0.7121$\pm$0.0120} & \textcolor{red}{0.8623$\pm$0.0069}  \\ 
		\midrule
		\multirow{5}{*}{100Leaves}
		& KM++ \cite{www/Sculley10}  & \textcolor{blue}{0.6134$\pm$0.0087} & \textcolor{blue}{0.8120$\pm$0.0038} & \textcolor{blue}{0.4914$\pm$0.0108} & 0.6841$\pm$0.0065 & \textcolor{blue}{0.5940$\pm$0.0091} \\
		& GAE \cite{corr/KipfW16a}        & 0.2875$\pm$0.0180 & 0.6545$\pm$0.0110 & 0.1772$\pm$0.0124 & 0.4525$\pm$0.0128 & 0.2664$\pm$0.0224 \\
		& DAEGC \cite{ijcai/WangPHLJZ19} & 0.5625$\pm$0.0000 & 0.7988$\pm$0.0003 & 0.3850$\pm$0.0004 & \textcolor{blue}{0.7869$\pm$0.0002} & 0.5161$\pm$0.0006 \\
		& SDCN \cite{www/Bo0SZL020}       & 0.3683$\pm$0.0465 & 0.6737$\pm$0.0301 & 0.2406$\pm$0.0424 & 0.4626$\pm$0.0489 & 0.3451$\pm$0.0485 \\
		& CMGEC      & \textcolor{red}{0.9156$\pm$0.0070} & \textcolor{red}{0.9684$\pm$0.0025} & \textcolor{red}{0.8876$\pm$0.0067} & \textcolor{red}{0.9461$\pm$0.0042} & \textcolor{red}{0.9086$\pm$0.0077} \\ 
		\midrule
		\multirow{5}{*}{Cub}  
		& KM++ \cite{www/Sculley10}  & 0.7243$\pm$0.0129 & 0.7085$\pm$0.0032 & 0.5543$\pm$0.0083 & 0.6992$\pm$0.0033 & 0.7325$\pm$0.0129 \\
		& GAE \cite{corr/KipfW16a}        & 0.7917$\pm$0.0325 & \textcolor{blue}{0.7904$\pm$0.0138} & 0.6936$\pm$0.0238 & \textcolor{blue}{0.7837$\pm$0.0143} & 0.7843$\pm$0.0343 \\
		& DAEGC \cite{ijcai/WangPHLJZ19} & 0.7467$\pm$0.0005 & 0.7328$\pm$0.0001 & 0.6125$\pm$0.0000 & 0.7337$\pm$0.0003 & 0.7345$\pm$0.0002\\
		& SDCN \cite{www/Bo0SZL020}       & \textcolor{blue}{0.8025$\pm$0.0415} & 0.7894$\pm$0.0211 & \textcolor{blue}{0.7045$\pm$0.0199} & 0.7749$\pm$0.0201 & \textcolor{blue}{0.7851$\pm$0.0254} \\
		& CMGEC   &  \textcolor{red}{0.8467$\pm$0.0041} & \textcolor{red}{0.7951$\pm$0.0059} & \textcolor{red}{0.7117$\pm$0.0064} & \textcolor{red}{0.7980$\pm$0.0061} & \textcolor{red}{0.8465$\pm$0.0043} \\ 
		\bottomrule
    	\end{tabular}}\label{tab:singlecomp1}
\end{table*}

\begin{table*}[!ht]
    \small
	\centering
	\caption{Clustering results between single view methods and CMGEC on multi-view data with common attribute graph.}
	\resizebox{12.5cm}{!}{
	\begin{tabular}{llccccc}
		\toprule
		Datasets & Methods & ACC & NMI  & ARI &AMI & F1  \\ \midrule
		\multirow{5}{*}{Cora}     
		& KM++ \cite{www/Sculley10} &  0.3311$\pm$0.0322 & 0.1302$\pm$0.0334 & 0.0597$\pm$0.0194 & 0.1267$\pm$0.0335 & 0.2504$\pm$0.0343 \\
		& GAE \cite{corr/KipfW16a} &  0.5301$\pm$0.0386 & 0.3971$\pm$0.0259 & 0.2933$\pm$0.0243 & 0.3875$\pm$0.0235 & 0.5019$\pm$0.0435 \\
		& DAEGC \cite{ijcai/WangPHLJZ19} &  \textcolor{blue}{0.6969$\pm$0.0002} & \textcolor{red}{0.5341$\pm$0.0004} & \textcolor{red}{0.4690$\pm$0.0001} & \textcolor{red}{0.5318$\pm$0.0005} & \textcolor{blue}{0.6839$\pm$0.0003} \\
		& SDCN \cite{www/Bo0SZL020}       &  0.6024$\pm$0.0043 & \textcolor{blue}{0.5004$\pm$0.0030} & 0.3902$\pm$0.0029 & \textcolor{blue}{0.4991$\pm$0.0035} & 0.6184$\pm$0.0044 \\
		& CMGEC    &  \textcolor{red}{0.7068$\pm$0.0304} & 0.4851$\pm$0.0184 & \textcolor{blue}{0.4172$\pm$0.0204} & 0.4806$\pm$0.0184 & \textcolor{red}{0.6967$\pm$0.0189}  \\ 
		\midrule
		\multirow{5}{*}{Citeseer}    
		& KM++ \cite{www/Sculley10}  &  0.4755$\pm$0.0584 & 0.2338$\pm$0.0457 & 0.2002$\pm$0.0461 & 0.2321$\pm$0.0458 & 0.4497$\pm$0.0582\\
		& GAE \cite{corr/KipfW16a} &  0.3802$\pm$0.0167 & 0.1746$\pm$0.0179 & 0.1613$\pm$0.0215 & 0.1825$\pm$0.0180 & 0.3633$\pm$0.0435 \\
		& DAEGC \cite{ijcai/WangPHLJZ19} &  0.6595$\pm$0.0001 & \textcolor{red}{0.4168$\pm$0.0000} & \textcolor{red}{0.4152$\pm$0.0000} & \textcolor{red}{0.4159$\pm$0.0001} & 0.6289$\pm$0.0000\\
		& SDCN \cite{www/Bo0SZL020}      &  \textcolor{blue}{0.6596$\pm$0.0031} & \textcolor{blue}{0.3871$\pm$0.0032} & 0.4017$\pm$0.0043 & \textcolor{blue}{0.3913$\pm$0.0041} & \textcolor{blue}{0.6362$\pm$0.0024}\\
		& CMGEC      &  \textcolor{red}{0.6765$\pm$0.0512} & 0.3666$\pm$0.0361 & \textcolor{blue}{0.4072$\pm$0.0361} & 0.3650$\pm$0.0361 & \textcolor{red}{0.6549$\pm$0.0522}\\ 
		\midrule
		\multirow{5}{*}{Pubmed}
		& KM++ \cite{www/Sculley10}  &  0.5989$\pm$0.0009 & \textcolor{blue}{0.3114$\pm$0.0031} & \textcolor{blue}{0.2814$\pm$0.0014} & \textcolor{blue}{0.3003$\pm$0.0040} & 0.5895$\pm$0.0004\\
		& GAE \cite{corr/KipfW16a}  &  0.6324$\pm$0.0167 & 0.2497$\pm$0.0259 & 0.2460$\pm$0.0268 & 0.2547$\pm$0.0260 & 0.6275$\pm$0.0179      \\
		& DAEGC \cite{ijcai/WangPHLJZ19} &  \textcolor{blue}{0.6712$\pm$0.0000} & 0.2663$\pm$0.0001 & 0.2782$\pm$0.0001 & 0.2621$\pm$0.0000 & \textcolor{blue}{0.6597$\pm$0.0002}\\
		& SDCN \cite{www/Bo0SZL020}       &  0.6578$\pm$0.0042 & 0.2947$\pm$0.0054 & 0.2546$\pm$0.0039 & 0.2959$\pm$0.0051 & 0.6516$\pm$0.0078\\
		& CMGEC     & \textcolor{red}{0.7055$\pm$0.0087} & \textcolor{red}{0.3428$\pm$0.0043} & \textcolor{red}{0.3345$\pm$0.0050} & \textcolor{red}{0.3427$\pm$0.0039} & \textcolor{red}{0.6966$\pm$0.0102} \\ 
		\bottomrule
    	\end{tabular}}\label{tab:singlecomp2}
\end{table*}

\begin{table*}[!ht]
    \small
	\centering
	\caption{Clustering results between single view methods and CMGEC on single-view data with multiply attribute graphs.}
	\resizebox{12.5cm}{!}{
	\begin{tabular}{llccccc}
		\toprule
		Datasets & Methods & ACC & NMI  & ARI &AMI & F1  \\ \midrule
		\multirow{5}{*}{DBLP}     
		& KM++ \cite{www/Sculley10} &  0.3864$\pm$0.0061 & 0.1153$\pm$0.0049 & 0.0671$\pm$0.0080 & 0.1145$\pm$0.0049 & 0.3195$\pm$0.0055 \\
		& GAE \cite{corr/KipfW16a}  &  0.5558$\pm$0.0139 & 0.3072$\pm$0.0073 & 0.2577$\pm$0.0061 & 0.3112$\pm$0.0080 & 0.5418$\pm$0.0124      \\
		& DAEGC \cite{ijcai/WangPHLJZ19}&  \textcolor{blue}{0.8733$\pm$0.0000} & \textcolor{blue}{0.6742$\pm$0.0000} & \textcolor{blue}{0.7014$\pm$0.0000} & \textcolor{blue}{0.6803$\pm$0.0000} & \textcolor{blue}{0.8617$\pm$0.0000} \\
		& SDCN \cite{www/Bo0SZL020}   &  0.6497$\pm$0.0039 & 0.2977$\pm$0.0018 & 0.3099$\pm$0.0033 & 0.2950$\pm$0.0041 & 0.6377$\pm$0.0029  \\
		& CMGEC  &  \textcolor{red}{0.9103$\pm$0.0039} & \textcolor{red}{0.7237$\pm$0.0021} & \textcolor{red}{0.7859$\pm$0.0062} & \textcolor{red}{0.7234$\pm$0.0030} & \textcolor{red}{0.9042$\pm$0.0042} \\ 
		\midrule
		\multirow{5}{*}{IMDB}    
		& KM++ \cite{www/Sculley10}  &  0.3154$\pm$0.0034 & 0.0119$\pm$0.0048 & 0.0028$\pm$0.0021 & 0.0109$\pm$0.0049 & 0.1799$\pm$0.0106\\
		& GAE \cite{corr/KipfW16a} &  \textcolor{blue}{0.4298$\pm$0.0134} & \textcolor{blue}{0.0402$\pm$0.0031} & \textcolor{blue}{0.0403$\pm$0.0019} & \textcolor{blue}{0.0398$\pm$0.0023} & \textcolor{blue}{0.4620$\pm$0.0141}       \\
		& DAEGC \cite{ijcai/WangPHLJZ19} &  0.3683$\pm$0.0013 & 0.0055$\pm$0.0004 & 0.0039$\pm$0.0001 & 0.0059$\pm$0.0003 & 0.3560$\pm$0.0009\\
		& SDCN \cite{www/Bo0SZL020}    & 0.4047$\pm$0.0030 & 0.0099$\pm$0.0009 & 0.0109$\pm$0.0011 & 0.0101$\pm$0.0008 & 0.3535$\pm$0.0029  \\
		& CMGEC  &  \textcolor{red}{0.4844$\pm$0.0123} & \textcolor{red}{0.0514$\pm$0.0091} & \textcolor{red}{0.0469$\pm$0.0077} & \textcolor{red}{0.0510$\pm$0.0080} & \textcolor{red}{0.5101$\pm$0.0201}  \\ 
		\midrule
		\multirow{5}{*}{ACM}
		& KM++ \cite{www/Sculley10}  &  0.6753$\pm$0.0113 & 0.3253$\pm$0.0047 & 0.3077$\pm$0.0106 & 0.3249$\pm$0.0047 & 0.6779$\pm$0.0116\\
		& GAE \cite{corr/KipfW16a}  &  0.6990$\pm$0.0161 & 0.4771$\pm$0.0083 & 0.4377$\pm$0.0070 & 0.4803$\pm$0.0090 & 0.7025$\pm$0.0156      \\
		& DAEGC \cite{ijcai/WangPHLJZ19} &  \textcolor{blue}{0.8909$\pm$0.0000} & \textcolor{blue}{0.6430$\pm$0.0000} & \textcolor{blue}{0.7046$\pm$0.0000} & \textcolor{blue}{0.6339$\pm$0.0000} & \textcolor{blue}{0.8906$\pm$0.0000} \\
		& SDCN \cite{www/Bo0SZL020}      &  0.8631$\pm$0.0052 & 0.5783$\pm$0.0088 & 0.6387$\pm$0.0110 & 0.5787$\pm$0.0080 & 0.8619$\pm$0.0060  \\
		& CMGEC     &  \textcolor{red}{0.9089$\pm$0.0073} & \textcolor{red}{0.6912$\pm$0.0036} & \textcolor{red}{0.7232$\pm$0.0106} & \textcolor{red}{0.6909$\pm$0.0057} & \textcolor{red}{0.9072$\pm$0.0059}\\
		\bottomrule
    	\end{tabular}}\label{tab:singlecomp3}
\end{table*}

\begin{table*}[!ht]
    \small
	\centering
	\caption{Performance comparisons of different multi-view methods on multi-view data without predefined attribute graph.}
	\resizebox{12.5cm}{!}{
	\begin{tabular}{llccccc}
		\toprule
		Datasets & Methods & ACC & NMI  & ARI & AMI & F1  \\ \midrule
		\multirow{7}{*}{3Sources}     
		& PMSC \cite{nn/KangZPZZPCX20}       & 0.4479$\pm$0.0939 & 0.1461$\pm$0.0583 & 0.1353$\pm$0.0959 & 0.1672$\pm$0.0540 & 0.4310$\pm$0.0394 \\ 
		& MCGC \cite{tip/ZhanNWY19}       & 0.5444$\pm$0.0000 & 0.4254$\pm$0.0000 & 0.4270$\pm$0.0000 & 0.4573$\pm$0.0000 & 0.5650$\pm$0.0000 \\
		& MVGL \cite{tcyb/ZhanZGW18}       & 0.4550$\pm$0.0000 & 0.4810$\pm$0.0000 & 0.4072$\pm$0.0000 & 0.4755$\pm$0.0000 & 0.4586$\pm$0.0000 \\
		& GMC \cite{tkde/WangYL20}        & \textcolor{blue}{0.6923$\pm$0.0000} & \textcolor{blue}{0.6216$\pm$0.0000} & \textcolor{blue}{0.4431$\pm$0.0000} & \textcolor{blue}{0.6044$\pm$0.0000} & \textcolor{blue}{0.6047$\pm$0.0000} \\ 
		& AE$^2$-NET \cite{cvpr/ZhangLF19} & 0.4929$\pm$0.0198 & 0.3884$\pm$0.0126 & 0.3268$\pm$0.0151 & 0.3399$\pm$0.0131 & 0.4348$\pm$0.0143 \\
		& RMSL \cite{iccv/LiZF0ZH19}       & 0.5219$\pm$0.0671 & 0.4840$\pm$0.0567 & 0.3912$\pm$0.0199 & 0.4560$\pm$0.0603 & 0.4851$\pm$0.0454 \\
		& CMGEC     & \textcolor{red}{0.7653$\pm$0.0307} & \textcolor{red}{0.6694$\pm$0.0143} & \textcolor{red}{0.6049$\pm$0.0446} & \textcolor{red}{0.6515$\pm$0.0147} & \textcolor{red}{0.6634$\pm$0.0355} \\ 
		\midrule
		\multirow{7}{*}{BBC}    
		& PMSC \cite{nn/KangZPZZPCX20}       & 0.6349$\pm$0.0014 & 0.3124$\pm$0.0020 & 0.3573$\pm$0.0014 & 0.3289$\pm$0.0082 & 0.3822$\pm$0.0007 \\
		& MCGC \cite{tip/ZhanNWY19}       & 0.6606$\pm$0.0000 & 0.3547$\pm$0.0000 & 0.3085$\pm$0.0000 & 0.4046$\pm$0.0000 & 0.3759$\pm$0.0000 \\
		& MVGL \cite{tcyb/ZhanZGW18}       & 0.6620$\pm$0.0000 & 0.3475$\pm$0.0000 & 0.3001$\pm$0.0000 & 0.3934$\pm$0.0000 & 0.3721$\pm$0.0000 \\
		& GMC \cite{tkde/WangYL20}        & 0.6891$\pm$0.0000 & 0.5577$\pm$0.0000 & 0.4745$\pm$0.0000 & 0.5611$\pm$0.0000 & 0.6306$\pm$0.0000 \\
		& AE$^2$-NET \cite{cvpr/ZhangLF19} & 0.7120$\pm$0.0136 & 0.4192$\pm$0.0044 & 0.4125$\pm$0.0038 & 0.4103$\pm$0.0041 & 0.6200$\pm$0.0197 \\
		& RMSL \cite{iccv/LiZF0ZH19}       & \textcolor{blue}{0.8365$\pm$0.0303} & \textcolor{blue}{0.6438$\pm$0.0350} & \textcolor{blue}{0.6816$\pm$0.0350} & \textcolor{blue}{0.6410$\pm$0.0353} & \textcolor{blue}{0.7239$\pm$0.0493} \\
		& CMGEC    & \textcolor{red}{0.8737$\pm$0.0061} & \textcolor{red}{0.7144$\pm$0.0119} & \textcolor{red}{0.7392$\pm$0.0115} & \textcolor{red}{0.7121$\pm$0.0120} & \textcolor{red}{0.8623$\pm$0.0069}  \\ 
		\midrule
		\multirow{7}{*}{100Leaves}
		& PMSC \cite{nn/KangZPZZPCX20}       & 0.5459$\pm$0.0165 & 0.4292$\pm$0.0253 & 0.4927$\pm$0.0159 & 0.4369$\pm$0.0164 & 0.3081$\pm$0.0153 \\ 
		& MCGC \cite{tip/ZhanNWY19}       & 0.7694$\pm$0.0000 & 0.8544$\pm$0.0000 & 0.4924$\pm$0.0000 & 0.7926$\pm$0.0000 & 0.4987$\pm$0.0000 \\
		& MVGL \cite{tcyb/ZhanZGW18}       & 0.8106$\pm$0.0000 & 0.8912$\pm$0.0000 & 0.5155$\pm$0.0000 & \textcolor{blue}{0.8557$\pm$0.0000} & 0.5217$\pm$0.0000 \\
		& GMC \cite{tkde/WangYL20}        & \textcolor{blue}{0.8238$\pm$0.0000} & \textcolor{blue}{0.9292$\pm$0.0000} & 0.4974$\pm$0.0000 & 0.8479$\pm$0.0000 & 0.5042$\pm$0.0000 \\ 
		& AE$^2$-NET \cite{cvpr/ZhangLF19} & 0.7500$\pm$0.0210 & 0.8880$\pm$0.0134 & \textcolor{blue}{0.6714$\pm$0.0316}& 0.8106$\pm$0.0224 & \textcolor{blue}{0.7288$\pm$0.0226} \\
		& RMSL \cite{iccv/LiZF0ZH19}       & 0.6483$\pm$0.0049 & 0.8047$\pm$0.0096 & 0.4904$\pm$0.0014 & 0.6683$\pm$0.0158 & 0.5176$\pm$0.0160 \\
		& CMGEC      & \textcolor{red}{0.9156$\pm$0.0070} & \textcolor{red}{0.9684$\pm$0.0025} & \textcolor{red}{0.8876$\pm$0.0067} & \textcolor{red}{0.9461$\pm$0.0042} & \textcolor{red}{0.9086$\pm$0.0077} \\ 
		\midrule
		\multirow{7}{*}{Cub}  
		& PMSC \cite{nn/KangZPZZPCX20}    & 0.7179$\pm$0.0014 & 0.7548$\pm$0.0031 & 0.6397$\pm$0.0009 & 0.7501$\pm$0.0026 & 0.6673$\pm$0.0006 \\ 
		& MCGC \cite{tip/ZhanNWY19}       & 0.7454$\pm$0.0000 & \textcolor{blue}{0.7959$\pm$0.0000} & 0.6499$\pm$0.0000 & 0.7842$\pm$0.0000 & 0.6790$\pm$0.0000 \\
		& MVGL \cite{tcyb/ZhanZGW18}      & 0.7491$\pm$0.0000 & \textcolor{red}{0.7972$\pm$0.0000} & \textcolor{blue}{0.6571$\pm$0.0000} & \textcolor{blue}{0.7891$\pm$0.0000} & 0.6853$\pm$0.0000 \\
		& GMC \cite{tkde/WangYL20}        & 0.7333$\pm$0.0000 & 0.7947$\pm$0.0000 & 0.6467$\pm$0.0000 & 0.7884$\pm$0.0000 & 0.6862$\pm$0.0000 \\ 
		& AE$^2$-NET \cite{cvpr/ZhangLF19} & \textcolor{blue}{0.7677$\pm$0.0292} & 0.7666$\pm$0.0255 & 0.6458$\pm$0.0445 & 0.7589$\pm$0.0264 & \textcolor{blue}{0.7518$\pm$0.0177} \\
		& RMSL \cite{iccv/LiZF0ZH19}       & 0.7423$\pm$0.0096 & 0.7231$\pm$0.0192 & 0.6072$\pm$0.0194 & 0.7142$\pm$0.0198 & 0.6484$\pm$0.0177 \\
		& CMGEC   &  \textcolor{red}{0.8467$\pm$0.0041} & 0.7951$\pm$0.0059 & \textcolor{red}{0.7117$\pm$0.0064} & \textcolor{red}{0.7980$\pm$0.0061} & \textcolor{red}{0.8465$\pm$0.0043} \\ 
		\bottomrule
    	\end{tabular}}\label{tab:multicomp1}
\end{table*}

\begin{table*}[!ht]
    \small
	\centering
	\caption{Clustering results of different multi-view methods on multi-view data with common attribute graph.}
	\resizebox{12.5cm}{!}{
	\begin{tabular}{llccccc}
		\toprule
		Datasets & Methods & ACC & NMI  & ARI &AMI & F1  \\ \midrule
		\multirow{4}{*}{Cora}     
		& MCGC \cite{tip/ZhanNWY19}       & 0.3043$\pm$0.0000 & 0.0038$\pm$0.0000 & 0.0131$\pm$0.0000 & 0.0040$\pm$0.0000 & 0.3030$\pm$0.0000 \\
		& MVGL \cite{tcyb/ZhanZGW18}       & 0.2371$\pm$0.0000 & 0.0631$\pm$0.0000 & 0.0266$\pm$0.0000 & 0.0599$\pm$0.0000 & 0.2574$\pm$0.0000 \\
		& GMC \cite{tkde/WangYL20}        & \textcolor{blue}{0.3667$\pm$0.0000} & \textcolor{blue}{0.1389$\pm$0.0000} & \textcolor{blue}{0.0301$\pm$0.0000} & \textcolor{blue}{0.1914$\pm$0.0000} & \textcolor{blue}{0.3182$\pm$0.0000} \\
		& CMGEC    &  \textcolor{red}{0.7068$\pm$0.0304} & \textcolor{red}{0.4851$\pm$0.0184} & \textcolor{red}{0.4172$\pm$0.0204} & \textcolor{red}{0.4806$\pm$0.0184} & \textcolor{red}{0.6967$\pm$0.0189}  \\ 
		\midrule
		\multirow{4}{*}{Citeseer}    
		& MCGC \cite{tip/ZhanNWY19}       & \textcolor{blue}{0.3204$\pm$0.0000} & \textcolor{blue}{0.1037$\pm$0.0286} & \textcolor{blue}{0.0286$\pm$0.0000} & \textcolor{blue}{0.1109$\pm$0.0000} & 0.2973$\pm$0.0000 \\
		& MVGL \cite{tcyb/ZhanZGW18}       & 0.2816$\pm$0.0000 & 0.0803$\pm$0.0000 & 0.0225$\pm$0.0000 & 0.0815$\pm$0.0000 & \textcolor{blue}{0.3043$\pm$0.0000} \\
		& GMC \cite{tkde/WangYL20}        & - & - & - & - & - \\
		& CMGEC      &  \textcolor{red}{0.6765$\pm$0.0512} & \textcolor{red}{0.3666$\pm$0.0361} & \textcolor{red}{0.4072$\pm$0.0361} & \textcolor{red}{0.3650$\pm$0.0361} & \textcolor{red}{0.6549$\pm$0.0522}\\ 
		\midrule
		\multirow{4}{*}{Pubmed}
		& MCGC \cite{tip/ZhanNWY19}       & \textcolor{blue}{0.4890$\pm$0.0000} & \textcolor{blue}{0.1251$\pm$0.0000} & \textcolor{blue}{0.1465$\pm$0.0000} & \textcolor{blue}{0.1210$\pm$0.0000} & 0.5060$\pm$0.0000 \\
		& MVGL \cite{tcyb/ZhanZGW18}       & 0.4604$\pm$0.0000 & 0.0463$\pm$0.0000 & 0.0094$\pm$0.0000 & 0.0501$\pm$0.0000 & 0.5039$\pm$0.0000 \\
		& GMC \cite{tkde/WangYL20}        & 0.4025$\pm$0.0000 & 0.0173$\pm$0.0000 & 0.0050$\pm$0.0000 & 0.0264$\pm$0.0000 & \textcolor{blue}{0.5203$\pm$0.0000} \\
		& CMGEC     & \textcolor{red}{0.7055$\pm$0.0087} & \textcolor{red}{0.3428$\pm$0.0043} & \textcolor{red}{0.3345$\pm$0.0050} & \textcolor{red}{0.3427$\pm$0.0039} & \textcolor{red}{0.6966$\pm$0.0102} \\ 
		\bottomrule
    	\end{tabular}}\label{tab:multicomp2}
\end{table*}

\begin{table*}[!ht]
    \small
	\centering
	\caption{Clustering results of multi-view method O2MAC and CMGEC on single-view data with multiply attribute graphs.}
	\resizebox{12.5cm}{!}{
	\begin{tabular}{llccccc}
		\toprule
		Datasets & Methods & ACC & NMI  & ARI &AMI & F1  \\ \midrule
		\multirow{2}{*}{DBLP}     
		& O2MAC \cite{www/FanWSLLW20} & 0.9012$\pm$0.0048 & 0.7250$\pm$0.0116 & 0.7806$\pm$0.0088  & 0.7267$\pm$0.0109 & 0.8981$\pm$0.0050 \\
		& CMGEC  &  0.9103$\pm$0.0039 & 0.7237$\pm$0.0021 & 0.7859$\pm$0.0062 & 0.7234$\pm$0.0030 & 0.9042$\pm$0.0042 \\ 
		\midrule
		\multirow{2}{*}{IMDB}    
		& O2MAC \cite{www/FanWSLLW20} &  0.4586$\pm$0.0280 & 0.0607$\pm$0.0311 & 0.0732$\pm$0.0254 & 0.0593$\pm$0.0296 & 0.4676$\pm$0.0444 \\
		& CMGEC  & 0.4844$\pm$0.0123 & 0.0514$\pm$0.0091 & 0.0469$\pm$0.0077 & 0.0510$\pm$0.0080 & 0.5101$\pm$0.0201  \\ 
		\midrule
		\multirow{2}{*}{ACM}
		& O2MAC \cite{www/FanWSLLW20} &  0.9039$\pm$0.0042 & 0.6909$\pm$0.0087 & 0.7410$\pm$0.0110 & 0.6935$\pm$0.0089 & 0.9061$\pm$0.0101 \\
		& CMGEC     &  0.9089$\pm$0.0073 & 0.6912$\pm$0.0036 & 0.7232$\pm$0.0106 & 0.6909$\pm$0.0057 & 0.9072$\pm$0.0059\\
		\bottomrule
    	\end{tabular}}\label{tab:multicomp3}
\end{table*}

\begin{figure*}[!t]
	\centering
	\subfigure[BBC-Raw data]{	
		\label{fig2:a} 
		\includegraphics[width=3.4cm]{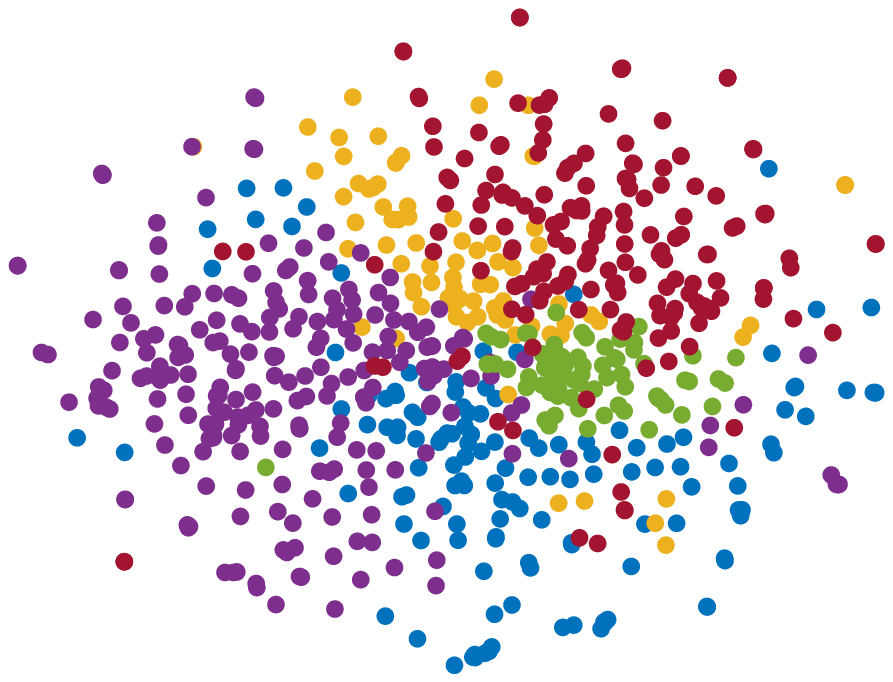}}
	\subfigure[BBC-GAE]{	
		\label{fig2:b}
		\includegraphics[width=3.4cm]{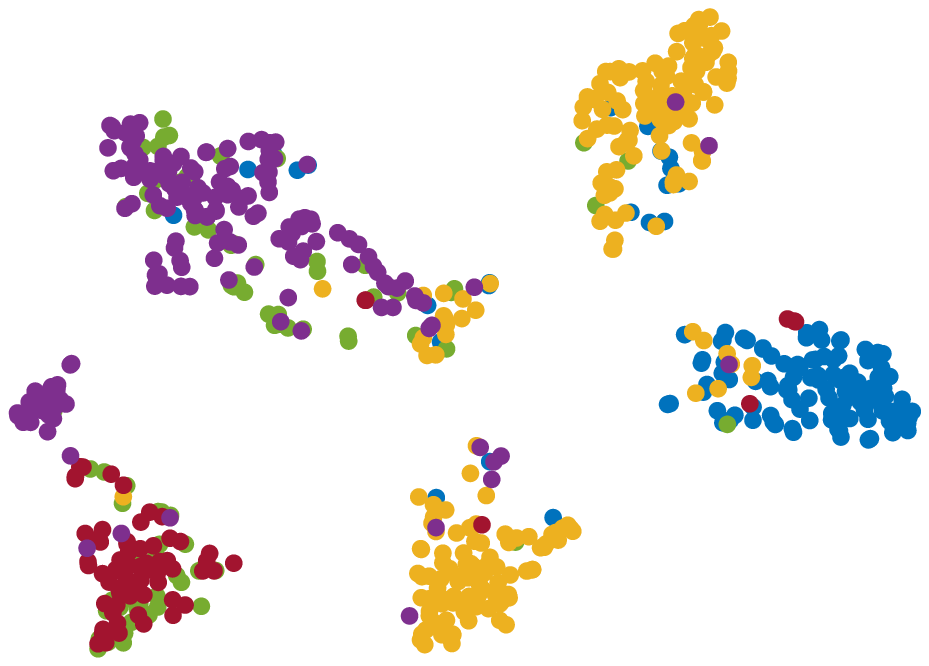}}
	\subfigure[BBC-DAEGC]{	
		\label{fig2:c} 
		\includegraphics[width=3.4cm]{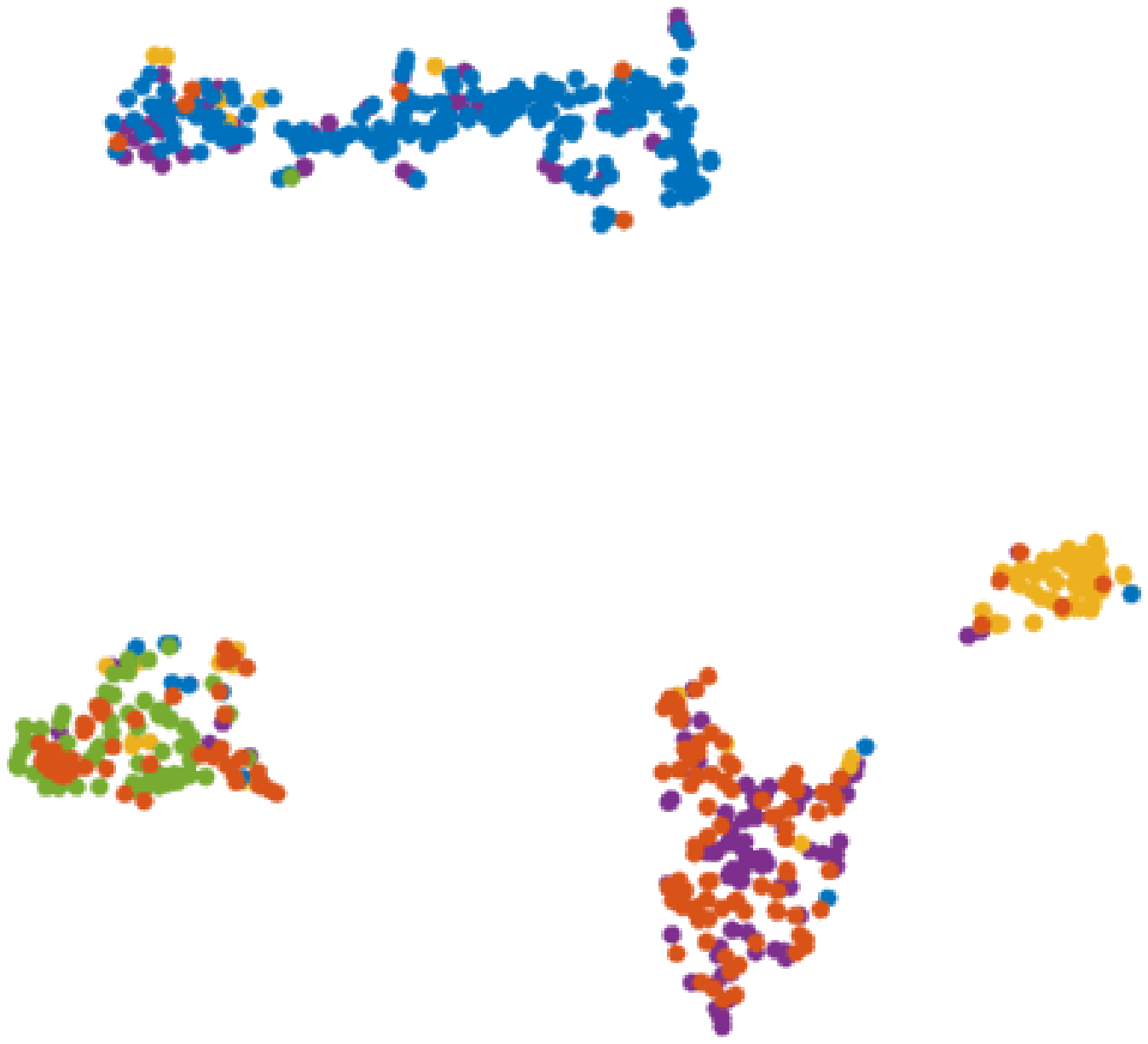}}
	\subfigure[BBC-RMSL]{	
		\label{fig2:d}
		\includegraphics[width=3.4cm]{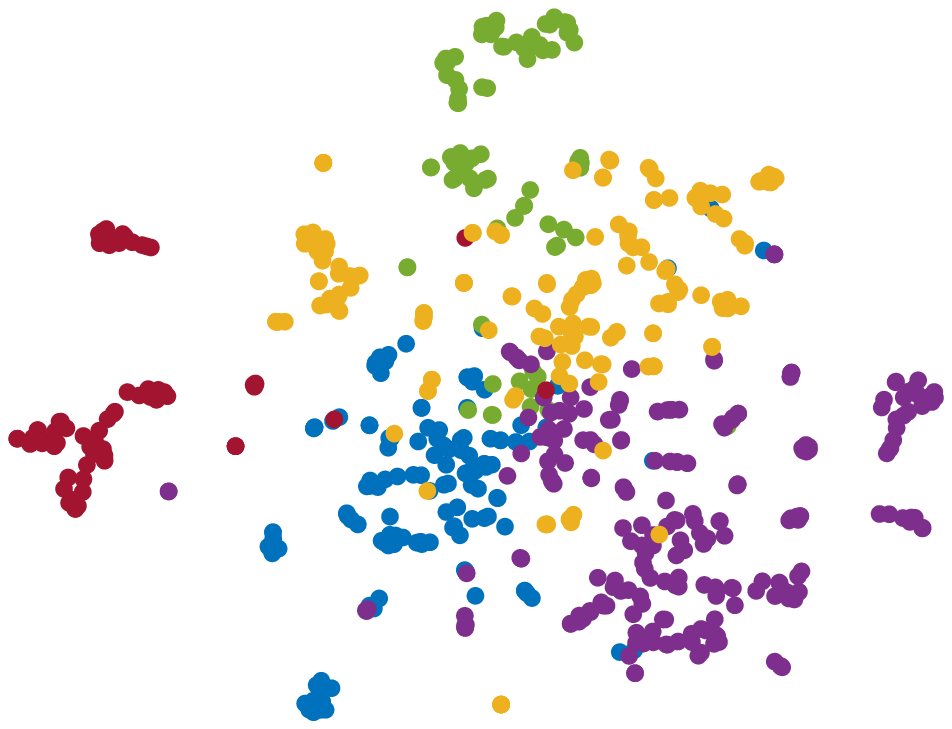}}
	\subfigure[BBC-CMGEC]{	
		\label{fig2:e}
		\includegraphics[width=3.4cm]{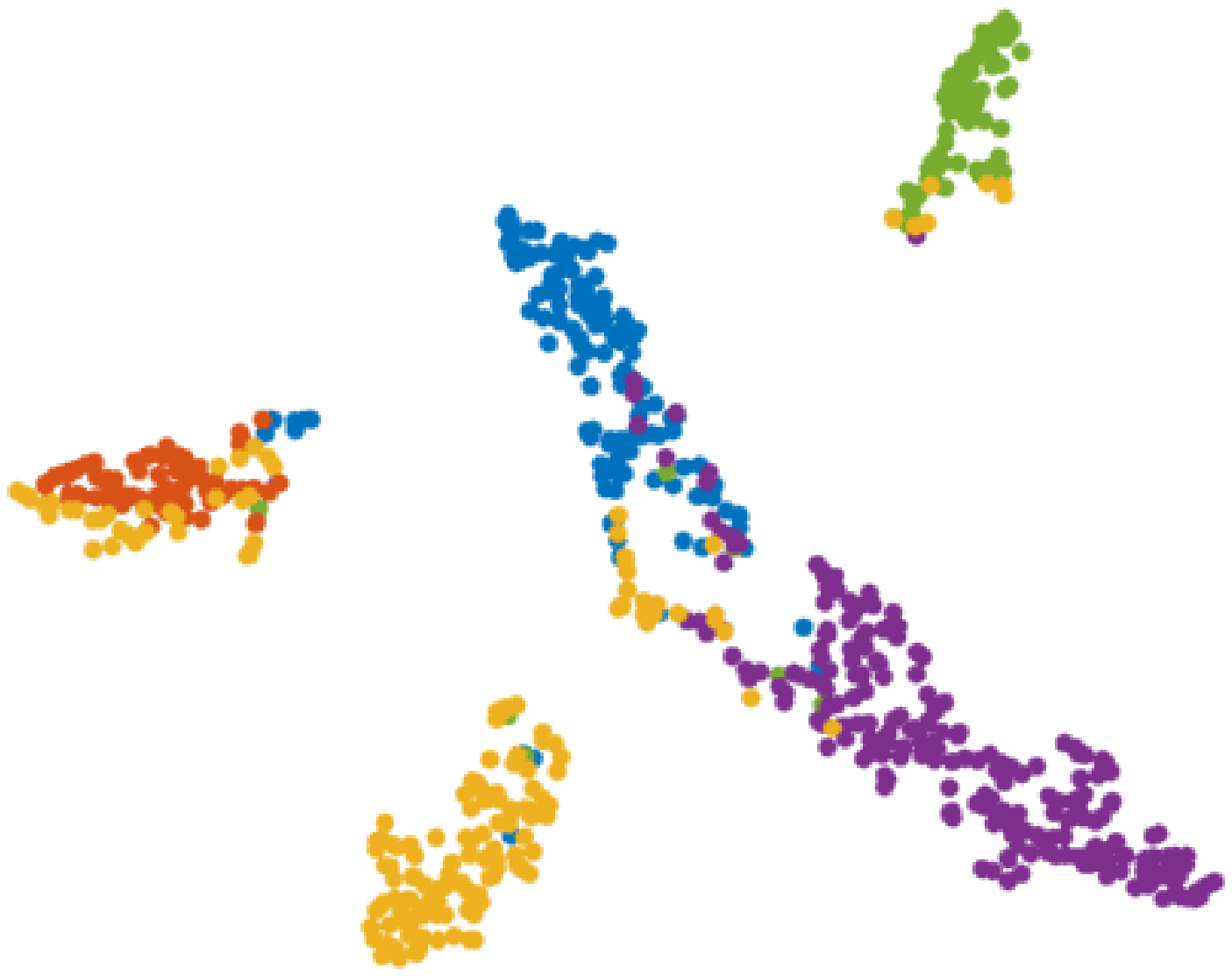}}
	\vfill
	\subfigure[3Sources-Raw data]{
		\label{fig2:f} 
		\includegraphics[width=3.4cm]{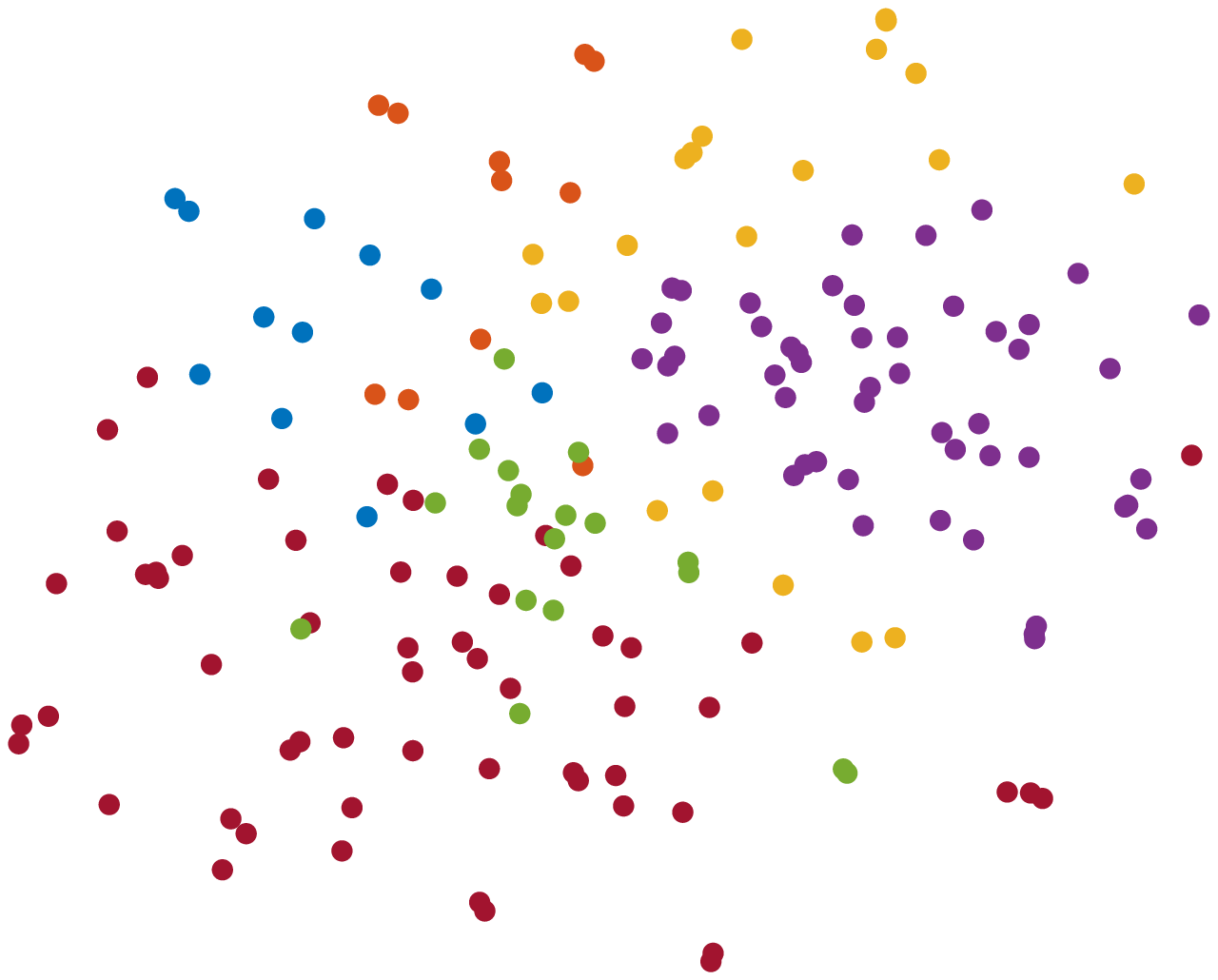}}
	\subfigure[3Sources-GAE]{
		\label{fig2:g} 
		\includegraphics[width=3.4cm]{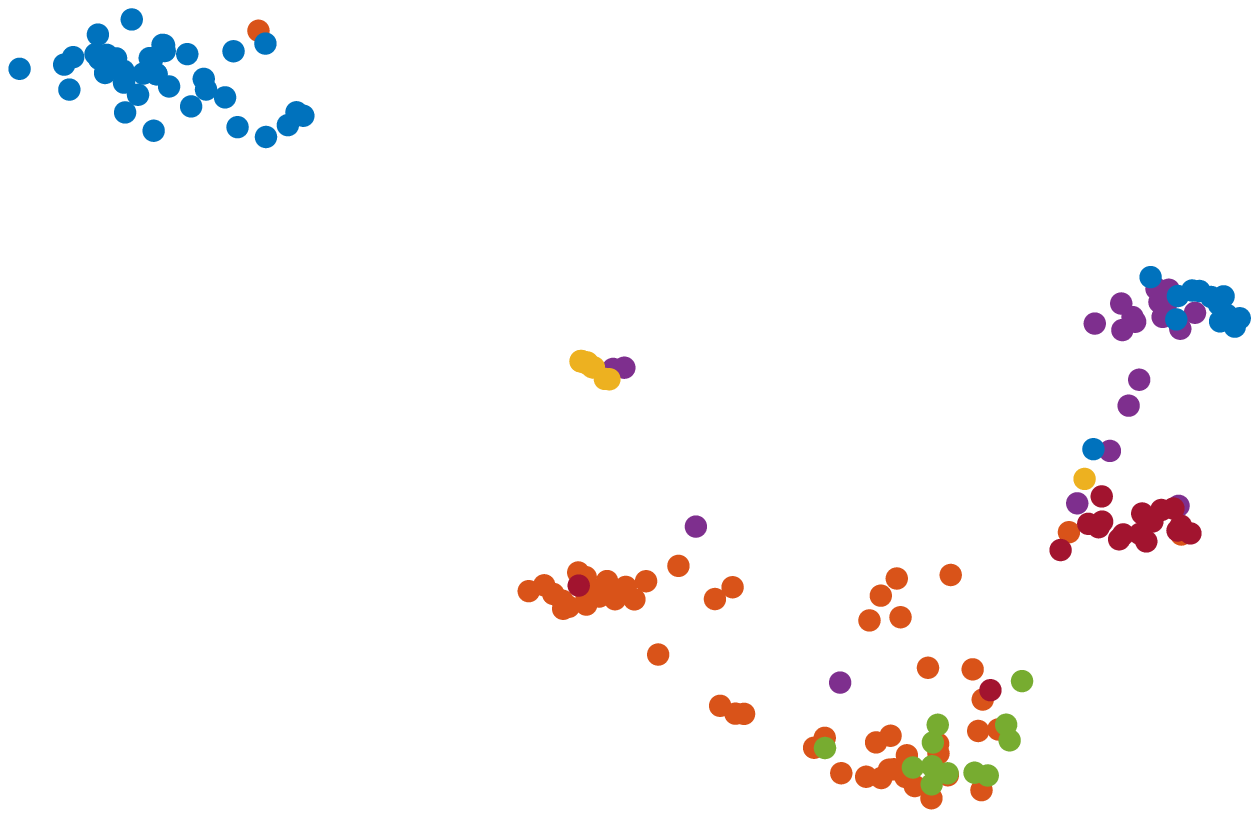}}
	\subfigure[3Sources-DAEGC]{	
		\label{fig2:h} 
		\includegraphics[width=3.4cm]{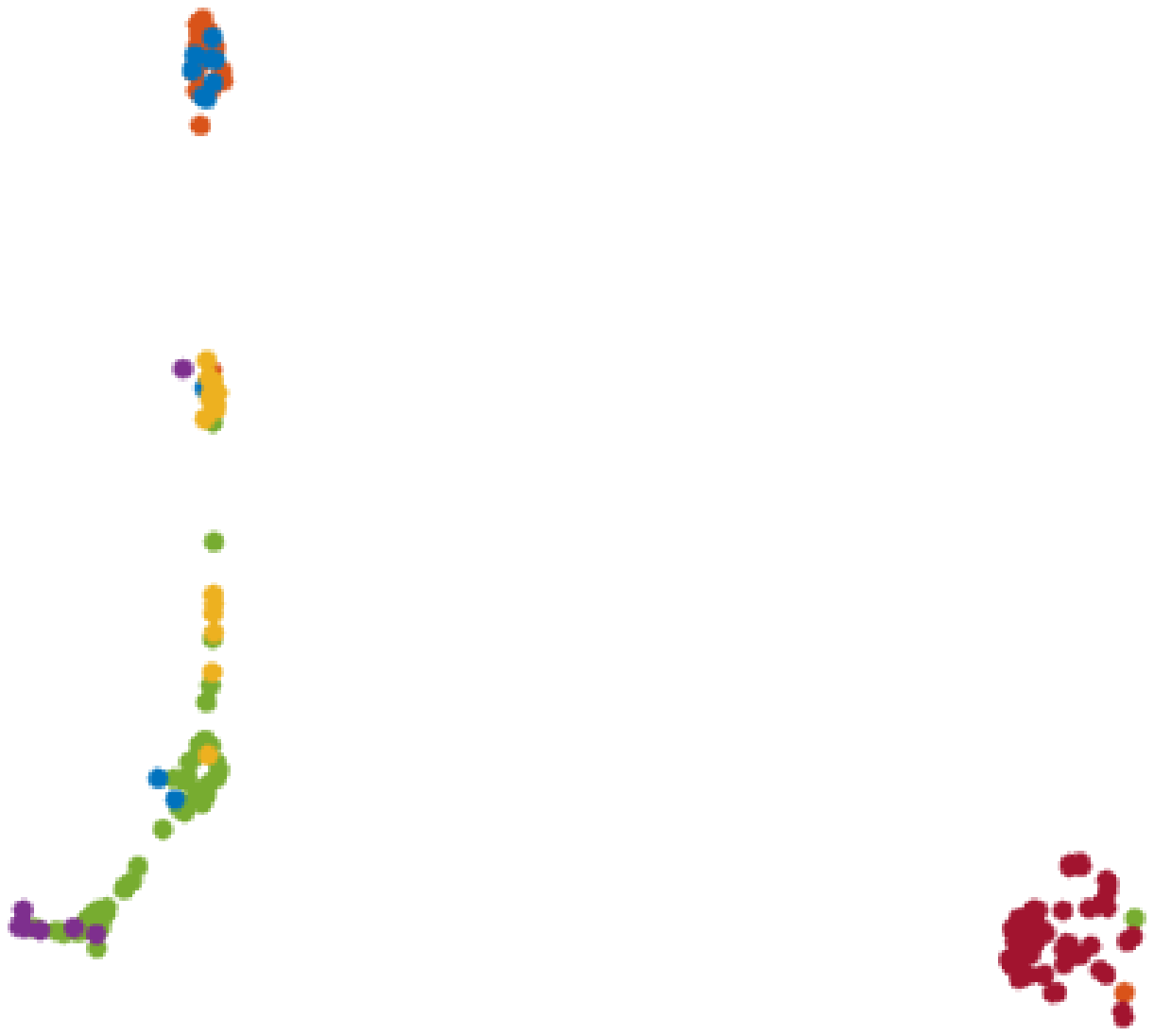}}
	\subfigure[3Sources-RMSL]{
		\label{fig2:i}
		\includegraphics[width=3.4cm]{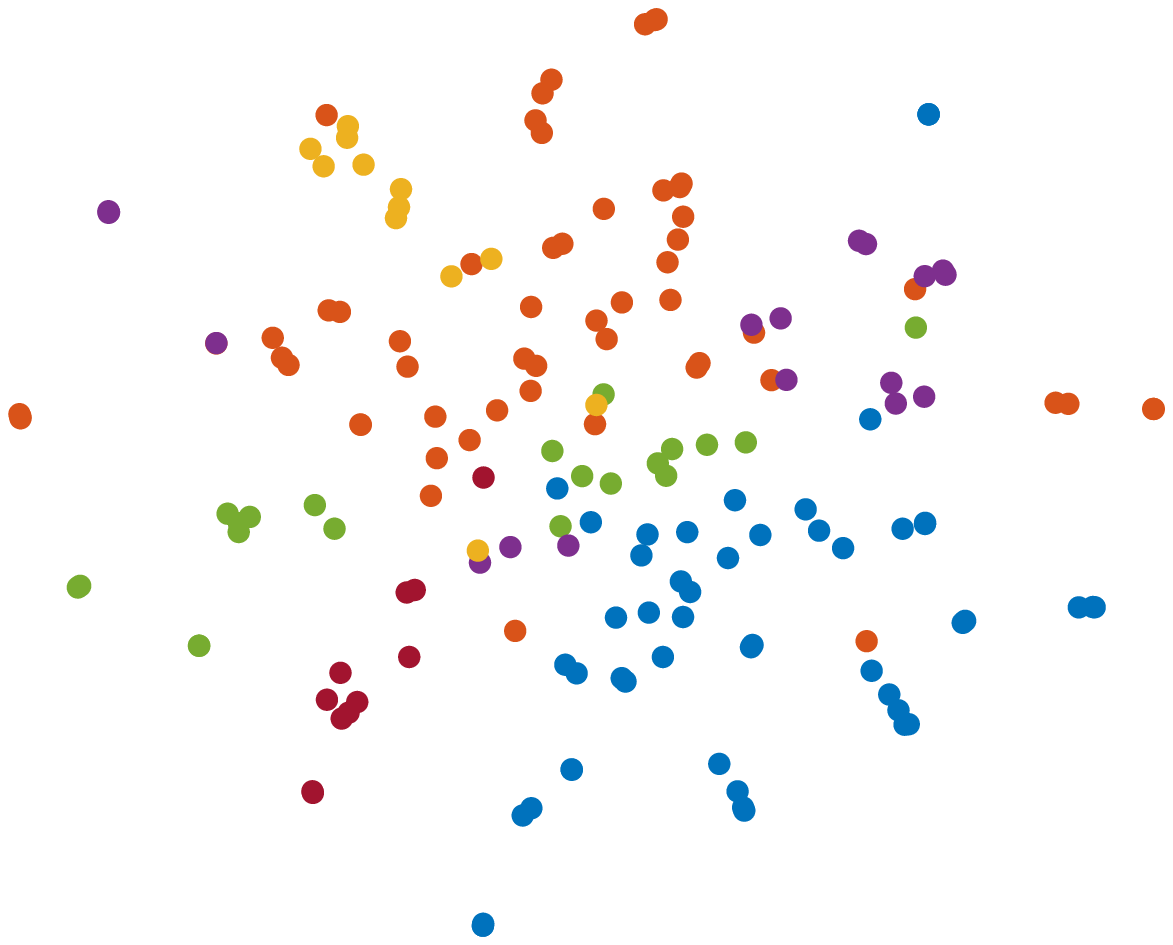}}
	\subfigure[3Sources-CMGEC]{
		\label{fig2:j}
		\includegraphics[width=3.4cm]{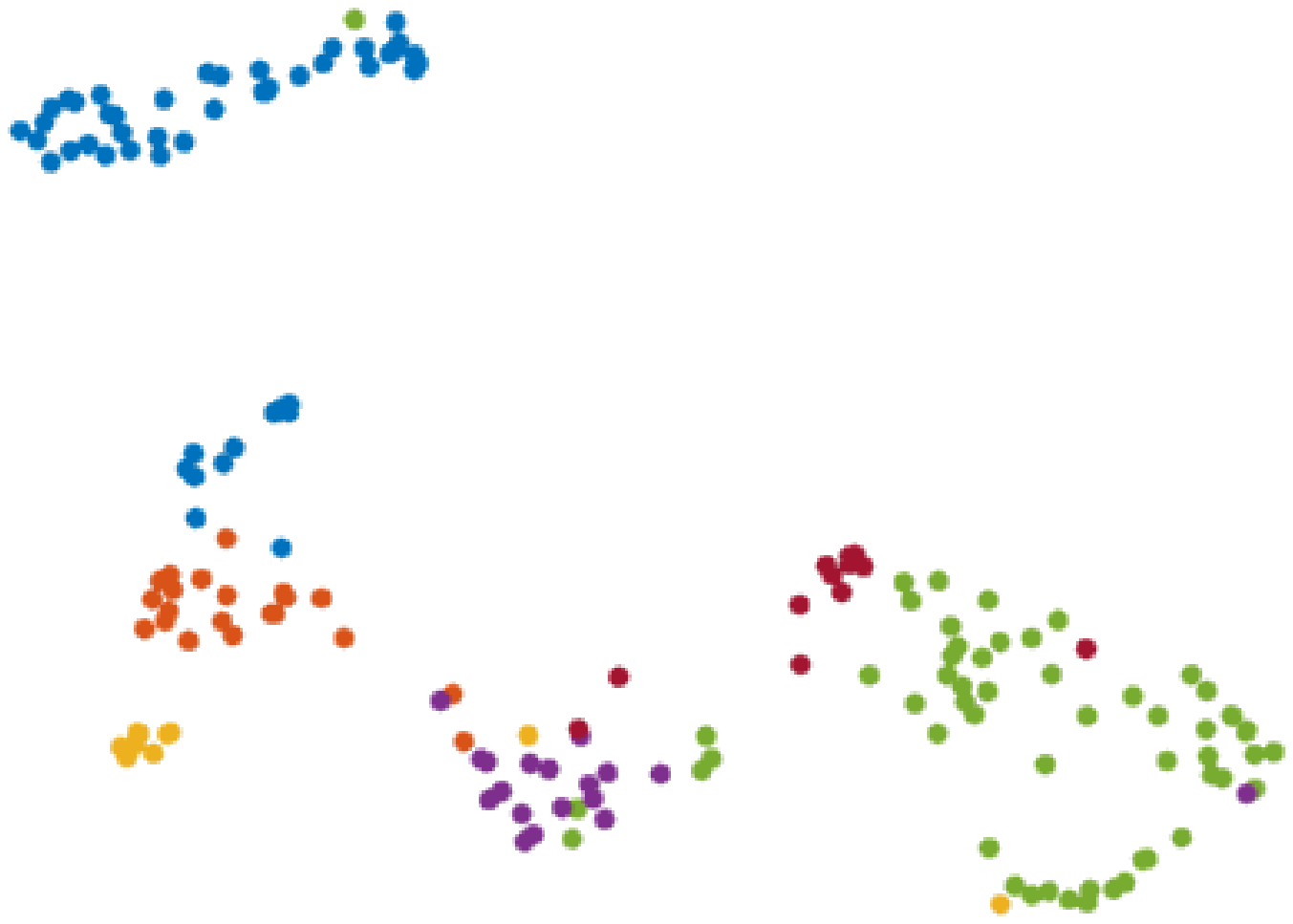}}
	\caption{t-SNE \cite{jmlr/laurens08} visualizations of representations learned by various methods on BBC (top row) and 3Sources (bottom row).}
	\label{fig:vis} 
\end{figure*}

\begin{table}[t]
    \tiny
	\centering
	\caption{Performance of different dimensions of the learned representation}
	\resizebox{8cm}{!}{
	\begin{tabular}{lcccc}
		\toprule
		Datasets & Dimensions & ACC & NMI  & ARI  \\ 
		\midrule
		\multirow{5}{*}{BBC}     
		& 5  &  0.8394 & 0.6849 & 0.6863 \\
        & 10 &  \textbf{0.8737} & \textbf{0.7144} & \textbf{0.7392} \\
        & 20 &  0.8458 & 0.6940 & 0.6965 \\
        & 30 &  0.8196 & 0.6879 & 0.6849 \\
        & 40 &  0.8394 & 0.6843 & 0.6850 \\
		\midrule
		\multirow{5}{*}{Cub}    
		& 5  &  0.7733 & 0.7453 & 0.6524 \\
        & 10 &  \textbf{0.8467} & 0.7951 & \textbf{0.7117} \\
        & 20 &  0.7956 & 0.7829 & 0.6795 \\
        & 30 &  0.8183 & \textbf{0.7974} & 0.6937 \\
        & 40 &  0.8161 & 0.7896	& 0.6916 \\
		\bottomrule
    	\end{tabular}}\label{tab:z}
\end{table}

\subsection{Experimental Settings}

\paragraph{Datasets} In order to fully evaluate the effectiveness of the proposed algorithm, we conduct experiments on three types of multi-view data: (a) Multi-view data without predefined attribute graph, including 3Source\footnote{http://mlg.ucd.ie/datasets/3sources.html}, BBC\footnote{http://mlg.ucd.ie/datasets/segment.html}, 100Leaves\footnote{https://archive.ics.uci.edu/ml/datasets/One-hundred+plant+species+ leaves+data+set}, and Cub\footnote{http://www.vision.caltech.edu/visipedia/CUB-200.html} \cite{WelinderEtal2010}; (b) Multi-view data with common attribute graph, including Cora, Citeseer and Pubmed\footnote{https://linqs.soe.ucsc.edu/data}; (c) Single-view data with multiply attribute graphs, including DBLP\footnote{https://dblp.uni-trier.de/}, IMDB\footnote{https://www.imdb.com/}, and ACM\footnote{http://dl.acm.org}. For convenience, these datasets are summarized in Table~\ref{tab:data1}, \ref{tab:data2} and \ref{tab:data3}.

\paragraph{Evaluation Metrics}For a comprehensive investigation, we evaluate the performance using five statistical metrics: Accuracy (ACC), Normalized Mutual Information (NMI), Adjusted Mutual Information (AMI), Adjusted Rand Index (ARI) and F1 measure(F1). Generally, the higher values of these five measures mean better clustering quality. 

\paragraph{Comparison Algorithms}

We compare the proposed CMGEC with some single-view clustering methods and several state-of-the-art multi-view clustering. 

\textbf{Single-view clustering methods:} K-means++ (KM++) \cite{www/Sculley10}, graph autoencoder (GAE) \cite{corr/KipfW16a}, deep attentional embedding graph clustering (DAEGC) \cite{ijcai/WangPHLJZ19}, and structural deep clustering network (SDCN) \cite{www/Bo0SZL020}. For the single view clustering methods, we report their results of the most informative view (achieves the best clustering performance).

\textbf{Multi-view clustering methods:} Partition level multiview subspace clustering (PMSC) \cite{nn/KangZPZZPCX20}, multiview consensus graph clustering (MCGC) \cite{tip/ZhanNWY19}, multiview graph learning (MVGL) \cite{tcyb/ZhanZGW18}, graph-based multi-view clustering(GMC) \cite{tkde/WangYL20}, autoencoder in autoencoder networks(AE$^2$-NET) \cite{cvpr/ZhangLF19}, reciprocal multi-layer subspace learning(RMSL) \cite{iccv/LiZF0ZH19}, and One2Multi graph autoencoder clustering framework(O2MAC) \cite{www/FanWSLLW20}.And we perform all algorithms 10 times and report the average results with the standard deviation.

\paragraph{Implementation Details}
In our experiments, we set $m = 10$, $\lambda_1= 0.01$, and $\lambda_2= 0.001$. For multi-view data without predefined graphs, we use $k$-NN algorithm to construct initial graphs. For multi-view data with a common attribute graph, the common graph is copied $V$ times and paired with the data from $V$ views to be fed into GFN and M-GAE. For single-view data with multiply attribute graphs, the single-view data is copied $V$ times and paired with the multiply attribute graphs from $V$ views to be fed into GFN and M-GAE. For each node, $k_M=3$ nearest neighbors are selected to compose positive pairs. Note that our model is not sensitive to the $k$ of initial $k$-NN graphs in a larger range. We set $k_G=10$ for all datasets. K-means++ \cite{www/Sculley10} is utilized to obtain the cluster results according to the learned common representation. All the experiments are conducted using the released code on an Ubuntu-18.04 OS with an NVIDIA RTX 3090 GPU. Some methods cannot perform on all types of data, and we only test them on partial datasets.

\subsection{Comparison of Clustering Performance}

\paragraph{Numerical Results Comparison}
The results of the proposed CMGEC compared with the single-view and multi-view methods on three types of public datasets are shown in Table~\ref{tab:singlecomp1}-\ref{tab:multicomp3}, respectively. The mean and variance of five used metrics are given, with the top value is highlighted in red font and the second-best in blue. Specifically, the performance comparisons of the single-view algorithms with our CMGEC are given in Table~\ref{tab:singlecomp1} to \ref{tab:singlecomp3}. From the results in these tables, we reach the following observations: (a) Generally speaking, compared with single-view baselines, the proposed CMGEC achieves better results on all datasets with most metrics, which shows the effectiveness of combining multi-view features for clustering; (b) GCN-based methods outperform the baseline KM++ in most cases, indicating that GCN can learn cluster-friendly embedding.

Moreover, the performance comparisons with multi-view algorithms are given in Table~\ref{tab:multicomp1} to \ref{tab:multicomp3}. The following observations can be made from the results: (a) Overall, CMGEC achieves very competitive and stable performance compared to almost all multi-view baselines. In many cases, the improvements are very significant. Taking the datasets BBC and 100Leaves for example, the ACC improvements of CMGEC over the second-best baseline are about 3.72\% and 9.18\%, respectively. The results demonstrate that the common representations learned by our proposed method are effective; (b) CMGEC consistently outperforms subspace-based methods PMSC and RMSL, indicating the effectiveness of learning node representations with the graph structure, because it can extract more inter-node information than only using node features. (c) Although for single-view data with multiple attribute maps, the performance of O2MAC is comparable to our approach. However, it is not easy to apply O2MAC directly to other types of multi-view data; (d) The performance of graph-based shallow methods is generally worse than GCN-based methods, confirming that it is useful to combine node characteristics with adjacent information.

\paragraph{Visualization of the clustering results}
In order to show the superiority of the representation obtained by our method, t-SNE \cite{jmlr/laurens08} is used to visualize the embedded feature space of different methods. And the visualizations on BBC and 3sources are given in Fig.~\ref{fig:vis}. From left to right, they are the space of raw data (best view), the results of GAE (best view), DAEGC (best view), RMSL, and our CMGEC, respectively. From Fig.~ \ref{fig:vis}, we can see that the representations obtained by our model are superior than that obtained by other algorithms which have clearer distribution structure.
 
\subsection{Parameter analysis}

\begin{figure}[!t]
	\centering
	\includegraphics[width=8.8cm]{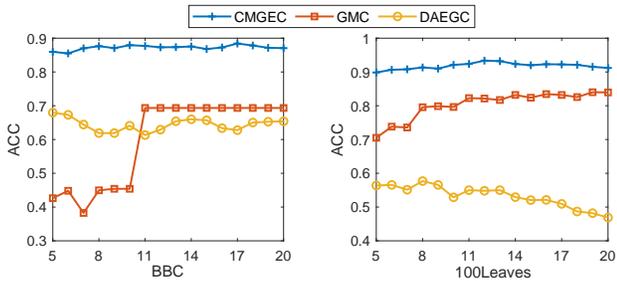}
	\caption{The parameter sensitivity of $k_G$ on BBC and 100Leaves datasets}
	\label{fig:kg} 
\end{figure}

\begin{figure}[!t]
	\centering
		\subfigure[ACC]{	
		\label{fig5:a}
		\includegraphics[width=4.2cm]{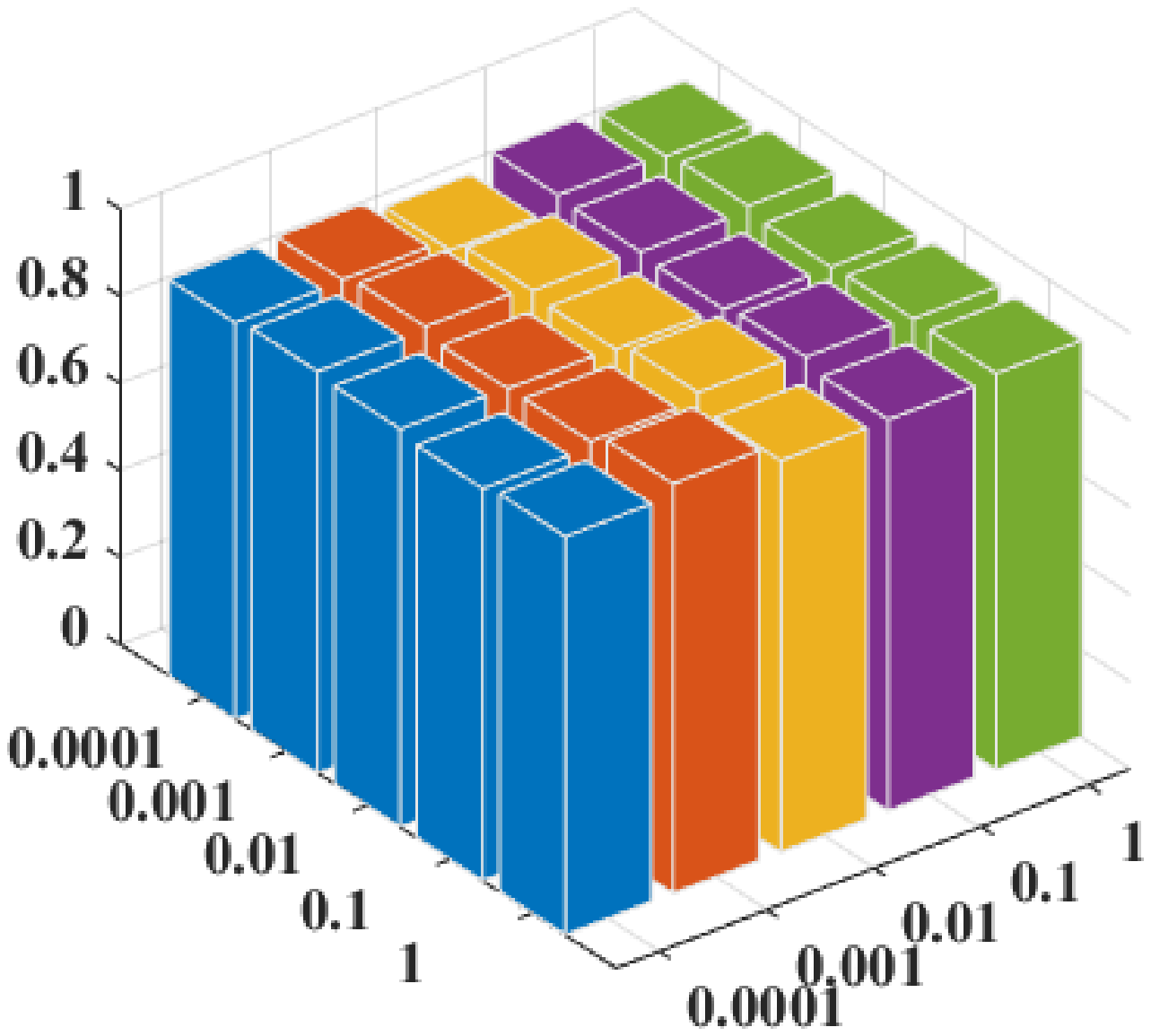}}
	\subfigure[NMI]{
		\label{fig5:b} 
		\includegraphics[width=4.2cm]{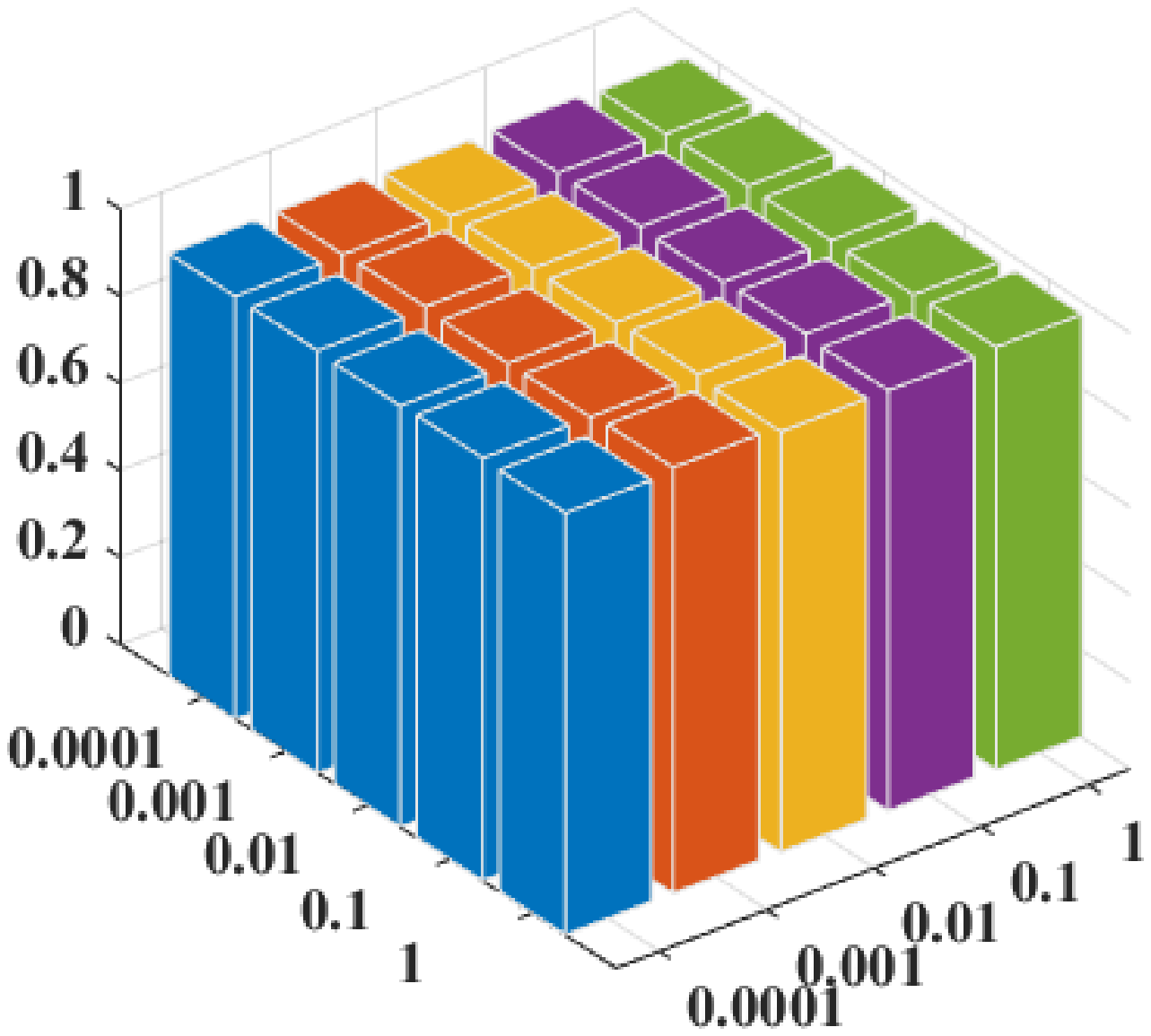}}
	\caption{The parameter effects of $\lambda_1$ and $\lambda_2$ on the BBC dataset with ACC and NMI metrics}
	\label{fig:lambda} 
\end{figure}

To better illustrate the stability of the proposed CMGEC, we perform experiments to analysis the sensitivity of the proposed method to the following parameters. 

\paragraph{The parameter sensitivity of $k_G$ in predefined graph}
The number of the nearest neighbors $k_G$ is an important parameter in the construction of the $k$-NN graph for data without predefined graph and has a great impact on the performance of most graph-based algorithms. To examine the effect of $k_G$, we design a $k_G$-sensitivity experiment on the BBC and 100Leaves datasets. It can be seen from Fig~\ref{fig:kg} that our model is insensitive with $k_G \in \{5,20\}$ compared with GMC and DAEGC. It proves that our method can learn multi-view structural information even there are less neighbor information or some spurious connections. However, larger $k_G$ can lead to more edges in the graph, which slows down the speed of the graph convolution. Thus, we set $k_G$ to 10 for all datasets in our experiments.

\paragraph{The parameter effect of $\lambda_1$ and $\lambda_2$}
In our CMGEC model, there are two hyperparameters $\lambda_1$ and $\lambda_2$ that need to be set properly. In our experiments, we tune $\lambda_1$ and $\lambda_2$ from $\{0.0001, 0.001, 0.01, 0.1, 1\}$ and $\{0.0001, 0.001, 0.01, 0.1, 1\}$, respectively. Fig~\ref{fig:lambda} shows the results of our method using different parameters (taking BBC as an example). Here, we vary a parameter at a time while keeping another fixed. From Fig~\ref{fig:lambda}, it can be seen that our method performs stably over a wide range of hyperparameter values.

\begin{figure}[!t]
	\centering
	\includegraphics[width=8cm]{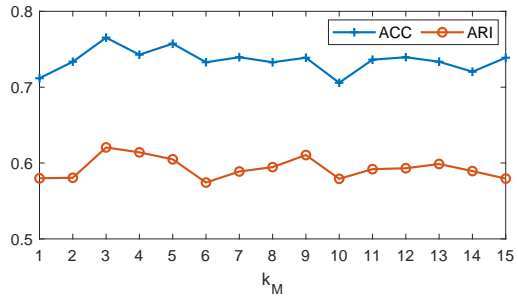}
	\caption{The parameter sensitivity of $k_M$ on 3Sources dataset}
	\label{fig:km} 
\end{figure}

\paragraph{The parameter analysis of $k_M$}
In MMIM, for each nodes, $k_M$ nearest neighbors are selected to compose positive pairs. In the following, we conduct experiments to show the effect of this parameter on the clustering performance. Fig~\ref{fig:km} presents the ACC and NMI of CMGEC by varying $k_M$ from 1 to 15. We can observe that the metrics first increases to a high value and generally maintains it up to slight variation with the increasing of $k_M$. CMGEC demonstrates stable performance across a wide range of $k_M$. For time-consuming reasons, we set $k_M$ to 3 in our experiments.

\paragraph{The sensitivity of the dimension of the common representation ($m$)} We vary the dimension of the learned representation from 5 to 40 and the results is given in Table~\ref{tab:z}. It can be observed that: when the dimension of representation changes from 5 to 10, the clustering performance improves significantly; however, when the dimension continue to increase, the clustering performance fluctuates, but the overall performance is still good. On the other hand, the time consumption of the algorithm increases with the dimensionality. Therefore, we set $m=10$ in our experiment.

\subsection{Ablation study}

\begin{figure*}[t]
	\centering
	\includegraphics[width=13cm]{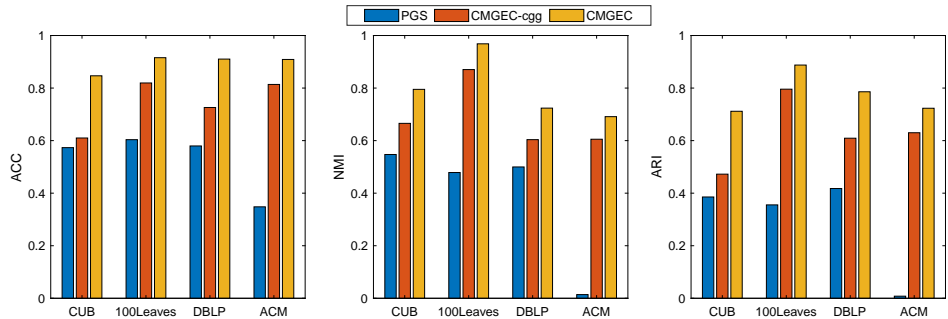}
	\vspace{-0.5em}
	\caption{Performance comparisons of PGS, CMGEC-cgg and CMGEC}
	\label{fig:cgg_cmgec}
\end{figure*} 

\begin{table*}[!t]
\tiny
\centering
\caption{The ablation study on three datasets. Different methods use modules identified by \CheckmarkBold. Best results are highlighted in \textbf{bold}.}
\resizebox{15cm}{!}{
\begin{tabular}{cccccccccccccccc}
\toprule
\multicolumn{1}{c}{\multirow{2}{*}{Method}} & \multirow{2}{*}{M-GAE} & \multirow{2}{*}{MMIM} & \multirow{2}{*}{GFN} & \multicolumn{3}{c}{3Sources} & \multicolumn{3}{c}{Cub}  & \multicolumn{3}{c}{BBC} & \multicolumn{3}{c}{ACM}\\ 
\multicolumn{1}{c}{}                        &                        &                       &                      & ACC      & NMI     & ARI     & ACC    & NMI    & ARI    & ACC    & NMI    & ARI    & ACC    & NMI    & ARI   \\
\midrule
CMGEC-MG                                    & \CheckmarkBold         &                       &                      & 0.6499   & 0.6360  & 0.5723  & 0.7699 & 0.7749 & 0.6487 & 0.7696 & 0.6113 & 0.6350 & 0.7483 & 0.5109 & 0.4427\\
CMGEC-G                                     & \CheckmarkBold         & \CheckmarkBold        &                      & 0.6568   & 0.6356  & 0.5967  & 0.7767 & 0.7740 & 0.6601 & 0.8091 & 0.6507 & 0.6643 & 0.7886 & 0.5709 & 0.4927 \\
CMGEC-M                                     & \CheckmarkBold         &                       & \CheckmarkBold       & 0.7459   & 0.6567  & 0.6027  & 0.8395 & 0.7860 & 0.6994 & 0.8610 & 0.6992 & 0.7254 & 0.8804 & 0.6851 & 0.6969 \\
CMGEC                                       & \CheckmarkBold         & \CheckmarkBold        & \CheckmarkBold       & \textbf{0.7653} & \textbf{0.6694} & \textbf{0.6049}  & \textbf{0.8467}  & \textbf{0.7951} & \textbf{0.7117} & \textbf{0.8737} & \textbf{0.7144} & \textbf{0.7392} & \textbf{0.9089} & \textbf{0.6912} & \textbf{0.7232}\\
\bottomrule
\end{tabular}}\label{tab:abl}
\end{table*}

In this section, the impact of each part on our CMGEC model are analyzed in detail. Specifically, we divided the ablation experiments into two parts according to how the clustering results are obtained as follows:

\paragraph{Clustering with consensus graph and predefined graph (Only with GFN)} Generally, the clustering results can be obtained from the representation or the graph segmentation \cite{pami/ShiM00}. In order to show the effectiveness of GFN module, we conduct comparison experiments on four datasets. The clustering results are given in Fig.~\ref{fig:cgg_cmgec}, where PGS denotes the clustering results obtained by the segmentation of the informative predefined graph, and CMGEC-cgg denotes the clustering results obtained by the segmentation of the consensus graph $A^*$ learned by GFN. From Fig.~\ref{fig:cgg_cmgec}, it can be seen that CMGEC-cgg performs better than PGS in all datasets, showing the effectiveness of learning consensus graph using GFN. Moreover, CMGEC outperforms PGS and CMGEC-cgg in all datasets, indicating that graph embedding can help to learn a suitable representation for clustering compared to consensus graph segmentation clustering.

\paragraph{Clustering without MMIM or GFN} 
To further investigate the effectiveness of diverse components of our model, we perform the following experiments to isolate the effect of GFN and MMIM. The clustering results are shown in Table~ \ref{tab:abl}, where CMGEC-MG (first row) means using M-GAE with $L_{rec}$ to obtain the common representation, CMGEC-G (second row) denotes M-GAE is trained using the whole loss $L_M$ but the graph of the most informative view is used as the consensus graph, CMGEC-M (third row) means M-GAE is trained using $L_{rec}$ and the consensus graph provided by GFN, and CMGEC (last row) means using all components to obtain the clustering results. From Table~ \ref{tab:abl}, we observe that: (a) It can be seen that each variants of our method has relatively high ACC, NMI, and ARI, and the best performance can be achieved when using whole CMGEC, which demonstrates that each part of the proposed model is significant for clustering task. (b) CMGEC-G generally achieves better cluster results than CMGEC-MG which suggests that mutual information contributes to the learning of more discriminative common representations; (c) We can clearly noticed that CMGEC-M significantly outperforms CMGEC-MG and CMGEC-G. It shows that learning a unified graph is essential to the learning of a suitable common representation and that using any view of the graph as the consensus graph is prejudiced.


\section{Conclusion}

In this paper, we propose a Consistent Multiple Graph Embedding Clustering framework (CMGEC), which is mainly composed of Multiple Graph Auto-Encoder (M-GAE), Multi-view Mutual Information Maximization module (MMIM) and Graph Fusion Network (GFN). Specifically, M-GAE is devised to learn a common representation using a multi-graph attention fusion encoder and reconstruct multi-view graphs by view-specific decoders. By introducing a multi-graph attention fusion layer, the common representation can adaptively integrate complementary information from multiple views. In order to maintain the similarity of the neighboring characteristic, MMIM is introduced to make similar instances still similar to each other in the common space. Moreover, we design a GFN to explore complex relationships among different views and learn a consensus graph needed in M-GAE. And the rank constraint on its Laplacian matrix is further utilized to train the GFN to improve the separability of the consensus graph. By jointly training these models, a view consistent representation can be learned for clustering. Experiments on three types of multi-view datasets verify the advantage of our proposed method compared with state-of-the-art methods.


%




\ifCLASSOPTIONcaptionsoff
  \newpage
\fi



\bibliographystyle{IEEEtran}
\bibliography{main}

\end{document}